%% file: main.tex
\input{constants}
\arxiv

\pdfoutput=1
\documentclass[10pt,twocolumn,letterpaper]{article}
\input{cvpr_header}
\begin{document}
\title{\paperTitle}
\author{\authorBlock}

\twocolumn[
    \maketitle
    \vspace{-3.0em}
    \input{figures/teaser}
    \bigbreak
]

\input{content/00_abstract}
\input{content/01_introduction}
\input{content/02_related_work}

\input{content/03_method}

\input{content/04_implementation}
\input{content/05_experiments}
\input{content/06_conclusion}

{\small
    \bibliographystyle{ieee_fullname}
    \bibliography{reference}
}

\ifarxiv \clearpage \input{content/09_supplementary} \fi

\end{document}

%% file: constants.tex
\def\cvprPaperID{0000} %
\def\confName{CVPR}
\def\confYear{2024}

\def\paperTitle{Relightable and Animatable Neural Avatar from Sparse-View Video}

\def\authorBlock{
    Zhen Xu $^{1}$ \qquad
    Sida Peng $^{1}$ \qquad
    Chen Geng $^{1}$ \qquad
    Linzhan Mou $^{1}$ \qquad \\
    Zihan Yan $^{2}$ \qquad
    Jiaming Sun $^{3}$ \qquad 
    Hujun Bao $^{1}$ \qquad
    Xiaowei Zhou $^{1}$ \\[0.5em]
    $^{1}$Zhejiang University \qquad
    $^{2}$MIT Media Lab \qquad
    $^{3}$Image Derivative Inc. \\
}

\newif\ifreview 
\newif\ifarxiv \newcommand{\arxiv}{\arxivtrue}
\newif\ifcamera 
\newif\ifrebuttal 

%% file: cvpr_header.tex
\ifreview \usepackage[review]{cvpr} \fi
\ifarxiv \usepackage[pagenumbers]{cvpr} \fi
\ifrebuttal \usepackage[rebuttal]{cvpr} \fi
\ifcamera \usepackage{cvpr} \fi

\usepackage{graphicx}
\usepackage{amsmath}
\usepackage{amssymb}
\usepackage{booktabs}

\input{macros}  %

\usepackage{xr-hyper}

\makeatletter
\newcommand*{\addFileDependency}[1]{
  \typeout{(#1)}
  \@addtofilelist{#1}
  \IfFileExists{#1}{}{\typeout{No file #1.}}
}

\makeatother

\usepackage[pagebackref,breaklinks,colorlinks]{hyperref}
\usepackage[capitalize]{cleveref}
\crefname{section}{Sec.}{Secs.}
\crefname{table}{Table}{Tables}
\crefname{figure}{Fig.}{Figs.}

%% file: macros.tex
\usepackage{times}
\usepackage{microtype}
\usepackage{epsfig}
\usepackage[table,xcdraw]{xcolor}
\usepackage{caption}
\usepackage{float}
\usepackage{placeins}
\usepackage{color, colortbl}
\usepackage{stfloats}
\usepackage{enumitem}
\usepackage{tabularx}
\usepackage{xstring}
\usepackage{multirow}
\usepackage{xspace}
\usepackage{url}
\usepackage{subcaption}
\usepackage{xcolor}
\usepackage{algorithm2e}
\usepackage[hang,flushmargin]{footmisc}

\ifcamera \usepackage[accsupp]{axessibility} \fi

\ifarxiv  \fi

\newcommand{\R}[1]{{%
    \textbf{%
        \ifstrequal{#1}{1}{\textcolor{red}{R#1}}{%
        \ifstrequal{#1}{2}{\textcolor{blue}{R#1}}{%
        \ifstrequal{#1}{3}{\textcolor{magenta}{R#1}}{%
        \ifstrequal{#1}{4}{\textcolor{teal}{R#1}}{%
                           \textcolor{cyan}{R#1}%
        }}}}%
    }%
}}

\usepackage{xcolor}
\definecolor{myred}{rgb}{0.8,0,0}
\definecolor{mygreen}{rgb}{0,0.8,0}

%% file: figures/teaser.tex
\begin{center}
    \includegraphics[width=0.99\textwidth]{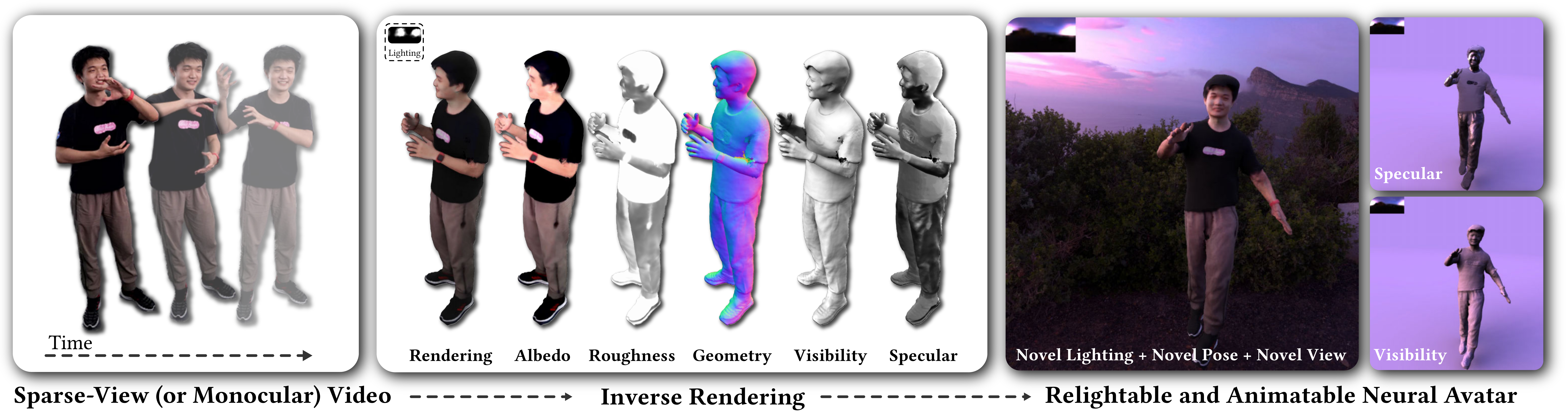}
    \captionof{figure}{
        \textbf{Reconstructing relightable and animatable neural avatar from sparse-view (or monocular) video.} Our method takes only a sparse-view (or monocular) video as input and reconstructs a relightable and animatable neural avatar under unknown illumination, which can then be relit with arbitrary environment lights and animated with arbitrary motion sequences. \textbf{Note that our method successfully captures the shininess of the skin and pants as well as the specular highlights on the t-shirt's plastisol printings.}
    }
    \label{fig:teaser}
\end{center}

%% file: content/00_abstract.tex
\begin{abstract}
This paper tackles the challenge of creating relightable and animatable neural avatars from sparse-view (or even monocular) videos of dynamic humans under unknown illumination.
Compared to studio environments, this setting is more practical and accessible but poses an extremely challenging ill-posed problem.
Previous neural human reconstruction methods are able to reconstruct animatable avatars from sparse views using deformed Signed Distance Fields (SDF) but cannot recover material parameters for relighting.
While differentiable inverse rendering-based methods have succeeded in material recovery of static objects, it is not straightforward to extend them to dynamic humans as it is computationally intensive to compute pixel-surface intersection and light visibility on deformed SDFs for inverse rendering. 
To solve this challenge, we propose a Hierarchical Distance Query (HDQ) algorithm to approximate the world space distances under arbitrary human poses.
Specifically, we estimate coarse distances based on a parametric human model and compute fine distances by exploiting the local deformation invariance of SDF.
Based on the HDQ algorithm, we leverage sphere tracing to efficiently estimate the surface intersection and light visibility. 
This allows us to develop the first system to recover animatable and relightable neural avatars from sparse view (or monocular) inputs.
Experiments demonstrate that our approach is able to produce superior results compared to state-of-the-art methods. Our code will be released for reproducibility.
\end{abstract}

%% file: content/01_introduction.tex
\section{Introduction}

Realistic human avatars have a range of applications \cite{starck2007surface,carranza2003free} in various domains, e.g., virtual reality, filmmaking, and video games.
This work focuses on the specific setting of creating animatable and relightable human avatars from sparse-view or monocular RGB videos.
This problem is challenging due to the inherent ambiguity of acquiring human geometry, materials, and motions from sparse view images \cite{debevec2000acquiring, peng2021neural}.
Traditional methods \cite{debevec2000acquiring,wenger2005performance,debevec2012light,guo2019relightables, starck2007surface, collet2015high, kanade1997virtualized} resolve this ambiguity via customized and costly capture devices, e.g., light stages with controllable illumination and dense camera arrays. However, such devices are restricted to professional users, impeding their universality and generalization.

Recent neural scene representation-based methods \cite{peng2022animatable, liu2021neural, wang2022arah} have demonstrated the ability to extract detailed geometry and photorealistic appearance of human performers from sparse-view videos without sophisticated studio setup.
These methods typically define the human model in canonical space and warp it into world space through a deformation module to represent human performers observed in videos.
For example, AniSDF \cite{peng2022animatable} models the human geometry and appearance as neural signed distance and radiance fields, and deforms them using linear blend skinning (LBS) \cite{lewis2000pose} and learned local deformation networks.
Albeit showing the capability of novel pose synthesis, the reconstructed avatars in these works \cite{peng2021animatable, liu2021neural, wang2022arah} are not relightable as they bake the shading and shadow into the appearance model. As a result, the shading of the avatars under novel poses is unrealistic and the environment illumination cannot be changed, which restricts the applicability of the avatars.

Another line of works attempts to create relightable models under natural illumination through inverse rendering techniques \cite{zhang2021physg,zhang2021nerfactor,srinivasan2021nerv, boss2021nerd,zhang2022modeling}, which estimate surface material parameters from input images through differentiable physically-based rendering.
Computing the visibility of 3D points to the environment light is essential for accurate estimation \cite{zhang2021nerfactor, zhang2022modeling}, but the cost of visibility computation is high.
To improve efficiency, L-Tracing \cite{chen2022tracing} adopts a signed distance field to represent the scene geometry and estimates the light visibility through sphere tracing, which iteratively marches along a ray using distance values until hitting the surface.
Although this strategy works well on static objects, it is not suitable for animatable neural avatars \cite{peng2022animatable, xu2021h, wang2022arah}, which warp the canonical SDF to world space based on a non-rigid motion field, producing a deformed SDF.
The reason is that sphere tracing might not converge on the deformed SDF \cite{seyb2019non} since the SDF is inherently defined in the canonical space, thereby yielding incorrect world-space distance.

In this work, we propose a novel approach for creating relightable and animatable human avatars from sparse-view (or monocular) videos via neural inverse rendering.
Inspired by previous methods \cite{peng2022animatable, wang2022arah}, we parameterize the human avatar as MLP networks, which predict material parameters and signed distance for any 3D point in canonical space.
These values are transformed into world space for rendering through a neural deformation field.
Our innovation lies in designing a hierarchical query scheme that enables a consistent approximation of 3D points' distance to the surface of the neural avatar under arbitrary human poses.
This allows us to perform sphere tracing for 3D points' pixel-surface intersection and light visibility for physically-based rendering.
Specifically,
we smoothly blend the world-space KNN (when query points are far from the surface) distances and canonical-space neural SDF (when query points are close to the surface), approximating an SDF defined on the world-space geometry of the neural avatar.
In this way, vanilla sphere tracing \cite{hart1993sphere} can be performed on the deformed SDF in world space when animating and relighting the avatar, avoiding the non-linearity of canonical sphere tracing, as well as the pitfalls of world space tracing with incorrect world-space distance.

Based on the Hierarchical Distance Query algorithm, we further develop a soft visibility computation scheme by incorporating traditional distance field soft shadow (DFSS) \cite{parker1998single} onto the deformed SDF, which is essential to the photorealism of the relightable neural avatar.
The soft shadow produced by an area light source typically requires multiple light samples to compute, while DFSS utilizes distance values to approximate the soft shadow coefficient with only a single sample.
Note that it is not trivial to combine DFSS with previous methods \cite{peng2021animatable,xu2021h,wang2022arah}, as they cannot produce world-space distance values from 3D points to the scene surface along an arbitrary direction.

To validate our approach, we collect a real-world multi-view dataset dubbed \textit{MobileStage}, which captures the complex shading and shadow effects of dynamic humans with an array of mobile phone cameras.
Furthermore, we extend the \textit{SyntheticHuman} dataset \cite{peng2022animatable} with novel illuminations, enabling the evaluation of relightable neural avatars with ground-truth photometric properties and relighting results.
Experiments on relighting ability and novel pose synthesis show that our method outperforms the state-of-the-art with superior visual quality and physical accuracy on both real-world and synthetic datasets. Our code will be made publicly available for reproducibility.

Our contributions can be summarized as follows: (a) We propose a novel system for reconstructing relightable and animatable neural avatars from sparse-view (or monocular) videos. (b) We design a hierarchical distance query algorithm for efficient pixel-surface intersection and light visibility computation using sphere tracing. (c) We extend DFSS to drivable neural SDF, efficiently producing realistic soft shadows for the neural avatars. (d) We demonstrate quantitative and qualitative improvements compared to prior work.

%% file: content/02_related_work.tex
\input{figures/pipeline}

\section{Related work}

\paragraph{Human avatars.}
To produce animatable human avatars, previous methods \cite{carranza2003free, starck2007surface,vlasic2008articulated,habermann2020deepcap,habermann2019livecap,xu2018monoperfcap} generally adopt a three-stage pipeline: they first reconstruct the human shape and appearance, then bind the shape to a skeleton, and finally animate the human model through linear blend skinning (LBS) algorithm \cite{lewis2000pose}.
Traditional methods tend to leverage complicated hardware, such as dense camera arrays \cite{starck2007surface, collet2015high, kanade1997virtualized, grau2003studio, starck2005virtual} or depth sensors \cite{aitpayev2012creation, shapiro2014rapid, tong2012scanning, bogo2015detailed}, to create high-fidelity human models.
Recently, some optimization-based methods \cite{alldieck2018video,peng2021neural,xu2021h,jiang2022neuman,weng2022humannerf} have attempted to reconstruct human models given sparse multi-view videos.
For example, Neural Body \cite{peng2021neural} represents a dynamic human model by combining SMPL model \cite{loper2015smpl} with neural radiance field (NeRF) \cite{mildenhall2021nerf}. Its model parameters are learned from images through differentiable volume rendering.

To animate the reconstructed human model, some \cite{huang2020arch, alldieck2018video} retrieve the skinning weights of the SMPL model for performing the LBS algorithm.
Several methods \cite{saito2021scanimate, peng2021animatable, huang2020arch, chen2021snarf} opt to optimize personalized skinning weights for the target human subject, where they represent the skinning weights as an MLP network and learn it from input data, such as human shapes \cite{saito2021scanimate, chen2021snarf} or multi-view videos \cite{peng2021animatable, huang2020arch}.
Another line of works \cite{peng2022animatable, liu2021neural} introduce a neural displacement field to improve animation realism.
The articulated deformation is represented by the LBS model of SMPL, and the non-rigid deformation is predicted using an MLP network.
While neural animatable methods can produce dynamic avatars that appear realistic, they do not model the material properties of the avatars, making them unable to adapt to different lighting conditions.

\paragraph{Relighting.}
To relight objects, a typical approach is first acquiring their material properties and then rendering with new illumination through physically-based rendering.
Traditional methods \cite{debevec2000acquiring,schmitt2020joint} mostly require a known illumination for calculating the material parameters through photometric stereos.
Light stage-based approaches \cite{debevec2000acquiring,wenger2005performance,debevec2012light,guo2019relightables} build a controllable light array to capture images of target subjects under multiple illuminations. Based on these captured images and the known illuminations, they solve for the unknown material properties.
Following this setting, some methods \cite{iwase2023relightablehands,li2023megane,bi2021deep} achieve photorealistic relighting results by adopting a neural renderer to implicit learn the relightable appearance model from light-stage images. However, these methods typically require the geometry to be known a priori.
More recently, neural inverse rendering methods \cite{zhang2021physg,zhang2021nerfactor,srinivasan2021nerv,boss2021nerd,chen2022tracing,zhang2022modeling,boss2022samurai,kuang2022neroic,boss2021neural} explore more flexible capture settings, where the illumination is unknown or even variable.
Motivated by its potential for many human-centric applications, research on human relighting has been widely conducted in the literature \cite{meka2020deep,pandey2021total,zhang2021neural,yeh2022learning}.
Same as other objects, the material properties of human subjects can be recovered using neural inverse rendering methods.
The difference is that human subjects exhibit more strong material priors.
Therefore, some methods \cite{pandey2021total,yeh2022learning,kanamori2019relighting,ji2022geometry,iqbal2022rana,alldieck2022photorealistic} attempt to train neural networks to predict human materials from a single image.
Recently, Relighting4D \cite{chen2022relighting4d} have attempted to acquire human materials from sparse multi-view videos.
However, Relighting4D is not designed to relight animatable avatars realistically, limiting its applicability.

%% file: figures/pipeline.tex
\begin{figure*}[ht]
    \centering
    \includegraphics[width=0.95\textwidth]{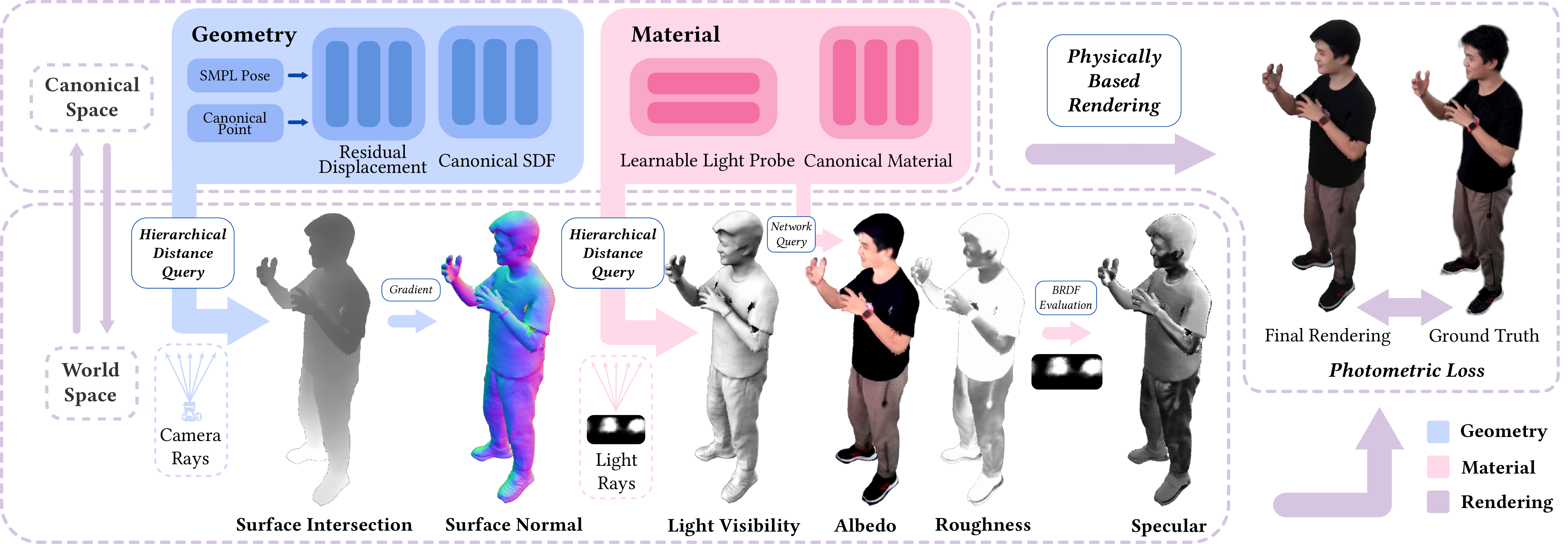}
    \caption{\textbf{Overview of the proposed approach.} Given world space camera rays, we perform sphere tracing on the hierarchically queried distances (Section~\ref{sec:hierarchical_distance_query}) to find surface intersections and canonical correspondences (Section~\ref{sec:geometry}). Light rays generated by an optimizable light probe are also sphere traced with HDQ to compute the closest distances along the ray for soft visibility (Section~\ref{sec:geometry}). Material properties (Section~\ref{sec:reflectance}) and surface normals are queried on the canonical correspondences and warped to world space. Then, the final pixel colors are computed using the rendering equation (Section~\ref{sec:training}).}
    \label{fig:pipeline}
\end{figure*}

%% file: content/03_method.tex
\section{Method}
\label{sec:method}

Given a sparse-view (or monocular) video of a human performer under natural and unknown illumination, we learn to reconstruct the drivable geometry and photometric properties of the human performer to create an animatable and relightable neural avatar.
We assume the human poses and the foreground masks are provided as in \cite{peng2022animatable,peng2021animatable,peng2021neural,liu2021neural}.

\subsection{Relightable and Animatable Neural Avatar}
\label{sec:avatar}

We formulate the relightable and animatable avatar using a set of canonical space neural fields and a warping between world and canonical space defined by the linear blend skinning algorithm \cite{lewis2000pose} and a displacement field \cite{peng2021animatable,peng2022animatable, liu2021neural, weng2022humannerf}.
In the canonical space, we define a set of geometry ($S(\boldsymbol{x})$) and material neural fields ($A(\boldsymbol{x})$ and $\Gamma(\boldsymbol{x})$) for the animated human model. The canonical space displacement field $F_{\Delta{\boldsymbol{x}}}$ provides additional pose-dependent deformation on top of SMPL inverse LBS.
More details about the warping process are provided in Section~\ref{sec:hierarchical_distance_query} and the supplementary.

The relightable and animatable neural avatar will be rendered by casting camera rays in world space and finding the surface intersection points $\boldsymbol{x}_s$ and their normals $\boldsymbol{n}_s$ using the Hierarchical Distance Query (HDQ) algorithm, whose material properties albedo $\alpha_s$ and roughness $\gamma_s$ can be obtained from the canonical mateiral MLPs $A(\boldsymbol{x})$ and $\Gamma(\boldsymbol{x})$, composing the BRDF $R_s(\boldsymbol{x}_s, \boldsymbol{\omega}_i, \boldsymbol{\omega}_o, \boldsymbol{n}_s)$. Light visibility $V_s(\boldsymbol{x}_s, \boldsymbol{\omega}_i)$ can be computed by performing HDQ sphere tracing on all incoming light directions.
We also incorporate Distance Field Soft Shadow (DFSS) algorithm \cite{ban2019area,aaltonen2018gpu,parker1998single} onto our drivable neural distance fields for soft-visibility computation.
These properties will be integrated around the hemisphere $\boldsymbol{\omega}_i \in \Omega$ using the rendering equation \cite{kajiya1986rendering}:
\begin{multline}
    \label{eq:the_rendering_equation}
    L_o = \\
    \int_{\Omega} L_s(\boldsymbol{\omega}_i) R_s(\boldsymbol{x}_s, \boldsymbol{\omega}_i, \boldsymbol{\omega}_o, \boldsymbol{n}_s) V_s(\boldsymbol{x}_s, \boldsymbol{\omega}_i) (\boldsymbol{n}_s \cdot \boldsymbol{\omega}_i) d\boldsymbol{\omega}_i,
\end{multline}
where $L_o(\boldsymbol{x_s}, \boldsymbol{\omega}_o) \in \mathbb{R}^3$ is the outgoing radiance at the surface intersection point $\boldsymbol{x}_s$, $\boldsymbol{\omega}_o$ is the outgoing radiance direction and $\boldsymbol{\omega}_i$ is the incoming radiance direction.
In this paper, we use the Microfacet BRDF model in \cite{walter2007microfacet} which is defined in the canonical space of the animatable avatar, and an optimizable light probe image $L_s(\boldsymbol{\omega}_i) \in \mathbb{R}^{16 \times 32 \times 3}$. An overview of the relightable and animatable avatar can be found in Figure~\ref{fig:pipeline}.

\input{figures/hdq}

\subsection{Hierarchical Distance Query}
\label{sec:hierarchical_distance_query}

Given the world space query point $\boldsymbol{x}$,
we approximate its world space distance $d^{world}(\boldsymbol{x})$ to the closest surface point on the neural avatar with the Hierarchical Distance Query algorithm $d^{world}(\boldsymbol{x}) \approx \tilde{d}^{world}(\boldsymbol{x}) = \mathrm{HDQ}(\boldsymbol{x})$, which is later used for Sphere Tracing \cite{hart1993sphere}.
The query algorithm consists of four stages: (a) coarse distance query, (b) inverse warping, (c) fine distance query, and (d) smooth distance blending.

\paragraph{Coarse distance query.}
\label{sec:coarse_distance_query}
We first perform a geodesically-aware signed $K$ Nearest Neighbor (GS-KNN) algorithm \cite{roussopoulos1995nearest} on the posed vertices $\boldsymbol{v} \in \mathcal{V}$ of the driven parametric human model (SMPL-H \cite{romero2017embodied}). GS-KNN produces the indices $\mathcal{I}_K= \{i_0,...,i_K\}$ of the $K$ closest points to $\boldsymbol{x}$ in $\mathcal{V}$, and its corresponding world-space closest vertices $\mathcal{V}_K=\{\boldsymbol{v}_0...,\boldsymbol{v}_K\}$, distances $\mathcal{D}_K=\{d_0...,d_K\}$, normals $\mathcal{N}_K=\{\boldsymbol{n}_0...,\boldsymbol{n}_K\}$ and blend weights $\mathcal{W}_K=\{\boldsymbol{w}_0...,\boldsymbol{w}_K\}$. We set $K=10$ through all experiments. The unsigned distance $\mathcal{D}$ is augmented with the sign of the dot product between $\boldsymbol{x}-\boldsymbol{v}$ and $\boldsymbol{n}$ to produce a coarse SDF. We additionally discard the $k$-th neighbor $\boldsymbol{v}_k$ if its canonical correspondence (T-Pose of SMPL-H) is too far from the canonical correspondence of the nearest neighbor, $K$ is set to 10 for all experiments.
The coarse level world space SDF is defined as $d_{coarse}^{world} = \frac{\sum_{k=0}^{K} d_k}{K}$.

\paragraph{Inverse warping.} We follow the previous literature\cite{peng2021animatable, liu2021neural} and use the linear blend skinning algorithm \cite{lewis2000pose} to perform the inverse warping. The details can be found in the supplementary material.

\paragraph{Fine distance query.}
\label{sec:fine_distance_query}
Given the warped query point $\boldsymbol{x}^\prime$, the pose-dependent displacement field $F_{\Delta{\boldsymbol{x}}}$ adds small perturbation to produce the final canonical space query point $\boldsymbol{x}^{\prime\prime}$.
We implement $F_{\Delta{\boldsymbol{x}}}$ as an MLP with the human pose at the $f$th frame $\Theta_f$ and $\boldsymbol{x}^\prime$ as input. The displaced canonical point $\boldsymbol{x}^{\prime\prime}$ fed into the canonical distance model $S$ is defined as
\begin{equation}
    \label{eq:canonical_space_main}
    \boldsymbol{x}^{\prime\prime} = \boldsymbol{x}^\prime + F_{\Delta{\boldsymbol{x}}}(\Theta_f, \boldsymbol{x}^\prime).
\end{equation}
Then, the fine canonical distance value can be obtained by querying the network $d_{fine}^{can} = S(\boldsymbol{x}^{\prime\prime})$.

\paragraph{Smooth distance blending.}
\label{sec:smooth_distance_blending}
Since SDF values of points close to the surface are hardly affected by LBS (Figure 2 and 5 of the supplementary), we propose to blend the fine canonical space distance value $d_{fine}^{can}$ and the coarse world space distance $d_{coarse}^{world}$ using a smooth function to produce the final approximated world space distance value $\tilde{d}^{world}$
\begin{multline}
    \tilde{d}^{world} = \\
    \begin{cases}
        d_{coarse}^{world}                                                                                & , \text{if } d_{coarse}^{world} > \tilde{T}_d \\
        d_{fine}^{can}  (1 - \frac{d_{fine}^{can}}{T_d}) + d_{coarse}^{world}  \frac{d_{fine}^{can}}{T_d} & , \text{otherwise}
    \end{cases}
\end{multline}
where $\tilde{T}_d$ is the distance threshold for cutting off coarse and fine distances, which is empirically set to $0.1$. Note that we only perform the evaluation of $S$ on points that satisfy the cutoff criteria $d_{coarse}^{world} \leq \tilde{T}_d$ for efficiency.

\subsection{Geometry}
\label{sec:geometry}

Our physically based renderer requires the pixel-surface intersection $\boldsymbol{x}_{s}\in \mathbb{R}^3$, surface normal $\boldsymbol{n}_{s} \in \mathbb{R}^3$, and light visibility $V(\boldsymbol{x}_{s}, \boldsymbol{\omega}_i) \in \mathbb{R}$ as input. Using the Hierarchical Distance Query, these values can be easily obtained from the world space SDF of the neural avatar under arbitrary human poses.

\paragraph{Surface intersection.}
Given a camera ray and the neural avatar's SDF, we compute the location $\boldsymbol{x}_s$ at which the ray $\boldsymbol{r}(t) = \boldsymbol{o} + t \boldsymbol{d}$ from the camera origin $\boldsymbol{o}$ along the ray direction $\boldsymbol{d}$ intersects the surface of the posed neural avatar.
Specifically, we perform $N_{st}$ Sphere Tracing iterations with the world space distance $\tilde{d}^{world} = \mathrm{HDQ}(\boldsymbol{x})$ using Hierarchical Distance Query until the ray converges to the surface intersection point $\boldsymbol{x}_s$.
The detailed algorithm is listed in the supplementary. $N_{st}$ is set to 16 across all experiments.

\paragraph{Surface normal.}
The analytic normal direction $\boldsymbol{n}$ of any 3D points could be computed as the gradient of the neural SDF using $\nabla \tilde{d}^{world}(\boldsymbol{x})$.
Although the hierarchical distance is differentiable, computing gradient through the whole query process is not efficient. Instead, we notice that surface intersections should satisfy the cutoff criteria of smooth distance blending in Section~{\ref{sec:hierarchical_distance_query}}, that is
\begin{equation}
    \tilde{d}^{world}(\boldsymbol{x}_{s}) = {d}^{can}_{fine}(\boldsymbol{x}_{s}), d_{coarse}^{world}(\boldsymbol{x}_{s}) \le \tilde{T}_d.
\end{equation}
Thus, the world space normal can be computed using $\nabla S(\boldsymbol{x}_{s}^{can})$ and transformed from canonical to world space using the forward warping process. More details can be found in the supplementary.

\paragraph{Light visibility.}
Light visibility $V(\boldsymbol{x}, \boldsymbol{\omega}_i)$ from any 3D point $\boldsymbol{x}$ along any light direction $\boldsymbol{\omega}_i$ can be computed as whether the light path $\boldsymbol{x} + t \boldsymbol{\omega}_i$ is occluded by the geometry of the posed neural avatar, which is later integrated in the rendering equation \cite{kajiya1986rendering} around the hemisphere. Since we use a discrete light probe $L_s(\boldsymbol{\omega}_i) \in \mathbb{R}^{16 \times 32 \times 3}$, the visibility term for every light direction needs to be integrated on the area of the pixel of $L_s(\boldsymbol{\omega}_i)$, which is time-consuming. Thanks to the global meaning of distance field, this occlusion and integration process can be approximated using Distance Field Soft Shadow (DFSS) \cite{parker1998single,ban2019area,aaltonen2018gpu}, producing soft visibility with a single light sample.
Specifically, we compute the visibility as the soft penumbra coefficient $p_{s}(\boldsymbol{x}_s, \boldsymbol{\omega}_i)$:
\begin{multline}
    \label{eq:soft_visibility}
    p_{s}(\boldsymbol{x}_s, \boldsymbol{\omega}_i) = \\
    \mathrm{min}(\frac{\tilde{d}^{world}(\boldsymbol{x}_s+t_{0}\boldsymbol{\omega}_i)}{2 t_{0} \sqrt{\frac{a}{\pi}}}, ..., \frac{\tilde{d}^{world}(\boldsymbol{x}_s+t_{N^{vis}_{st}}\boldsymbol{\omega}_i)}{2 t_{N^{vis}_{st}} \sqrt{\frac{a}{\pi}}}),
\end{multline}
for each surface point $\boldsymbol{x}_s$ along one of the 512 light directions $\boldsymbol{\omega}_i$ defined by $L_s(\boldsymbol{\omega}_i)$ during the $N^{vis}_{st}$ sphere tracing steps, which is set to 4 for all experiments.
The ratio between the two tangent values $\frac{\tilde{d}^{world}}{t}$ and $2\sqrt{\frac{a}{\pi}}$ serves as an approximation of the ratio of light being occluded by the geometry from $\boldsymbol{x}_s$ along $\boldsymbol{\omega}_i$.
Thanks to the smooth blending of $d^{world}_{coarse}$ and $d^{can}_{fine}$ in Section~\ref{sec:hierarchical_distance_query}, our soft visibility scheme produces realistic and smooth soft shadow even when distance values from the parametric human model \cite{romero2017embodied} and the canonical neural SDF are not perfectly aligned.
A detailed listing of this algorithm is provided in the supplementary.

\subsection{Reflectance}
\label{sec:reflectance}

We adopt the Microfacet BRDF model in \cite{walter2007microfacet} for our material representation, which is composed of a diffuse albedo $\alpha \in \mathbb{R}^3$ term and a specular roughness $\gamma \in \mathbb{R}$ term. We use a fixed Fresnel term of $0.04$.
Similar to \cite{zhang2021nerfactor,chen2022relighting4d,zhang2022modeling}, we paramterize the albedo and roughness map with two MLPs $\alpha = A(\boldsymbol{x}^{\prime\prime})$ and $\gamma = \Gamma(\boldsymbol{x}^{\prime\prime})$, which is defined in the same canonical frame as $S(\boldsymbol{x}^{\prime\prime})$ and $F_{\Delta}(\boldsymbol{x}^{\prime})$ in Section~\ref{sec:hierarchical_distance_query}.
The BRDF model is denoted $R_s(\boldsymbol{x}_s, \boldsymbol{\omega}_i, \boldsymbol{\omega}_o, \boldsymbol{n}_s)$ where $\boldsymbol{\omega}_i$ is the incoming radiance direction, $\boldsymbol{\omega}_o$ is the outgoing radiance direction and $\boldsymbol{n}_s$ is the surface normal.

Given world space query point $\boldsymbol{x}_s$ and its corresponding canonical space point $\boldsymbol{x}^{\prime\prime}$, we obtain the albedo $\alpha$ and roughness $\gamma$ by querying their canonical neural fields $A$ and $\Gamma$, which can then be converted to BRDF values as defined in \cite{walter2007microfacet}.
Our physically-based renderer also takes a light probe $L_s(\boldsymbol{\omega}_i) \in \mathbb{R}^{16 \times 32 \times 3}$ as input, which is represented by an optimizable neural texture during training and replaced with the designated environment map during relighting \cite{zhang2021nerfactor,chen2022relighting4d,debevec2008rendering}.

\subsection{Training}
\label{sec:training}

We use 512 discrete incoming light directions defined by the light probe $L_s(\boldsymbol{\omega}_i) \in \mathbb{R}^{16 \times 32 \times 3}$ to approximate the Rendering Equation \cite{kajiya1986rendering} as
\begin{multline}
    \label{eq:the_discrete_rendering_equation}
    L_o = \\
    \sum_{\boldsymbol{\omega}_i} L_s(\boldsymbol{\omega}_i) R_s(\boldsymbol{x}_s, \boldsymbol{\omega}_i, \boldsymbol{\omega}_o, \boldsymbol{n}_s) V_s(\boldsymbol{x}_s, \boldsymbol{\omega}_i) (\boldsymbol{n}_s \cdot \boldsymbol{\omega}_i) \Delta\boldsymbol{\omega}_i,
\end{multline}
where $\Delta \boldsymbol{\omega}_i$ is the solid angle of the incoming light $\boldsymbol{\omega}_i$ sampled from the light probe $L_s(\boldsymbol{\omega}_i)$ and $L_o(\boldsymbol{x_s}, \boldsymbol{\omega}_o) \in \mathbb{R}^3$ is the outgoing radiance at the surface intersection $\boldsymbol{x}_s$.

We optimize our relightable and animatable neural human avatar by rendering the image with given camera poses and comparing the pixel values $L_o$
against the ground truth ones $L_{gt}$. The main loss function is defined as $\mathcal{L}_{data} =\sum_{\boldsymbol{r} \in \mathcal{R}}  {\lVert L_o(\boldsymbol{r})- L_{gt}(\boldsymbol{r}) \rVert}_2$, where $\boldsymbol{r} = \boldsymbol{o} + t \boldsymbol{d} \in \mathcal{R}$ denotes all camera rays in the forward rendering process. Due to the ill-posed nature of the problem, we adopt a two-stage training strategy and add additional regularizations one the geometry (eikonal loss $\mathcal{L}_{eik}$) and material (sparsity loss $\mathcal{L}_{ent}$ and smoothness loss $\mathcal{L}_{a},\mathcal{L}_{r}$). We elaborate on the details of each loss term and the training stragety in the supplementary.
The training takes 20 hours on an Nvidia RTX 3090. Rendering a 512$\times$512 image takes 5s.

%% file: figures/hdq.tex
\begin{figure}
    \centering
    \includegraphics[width=0.40\textwidth]{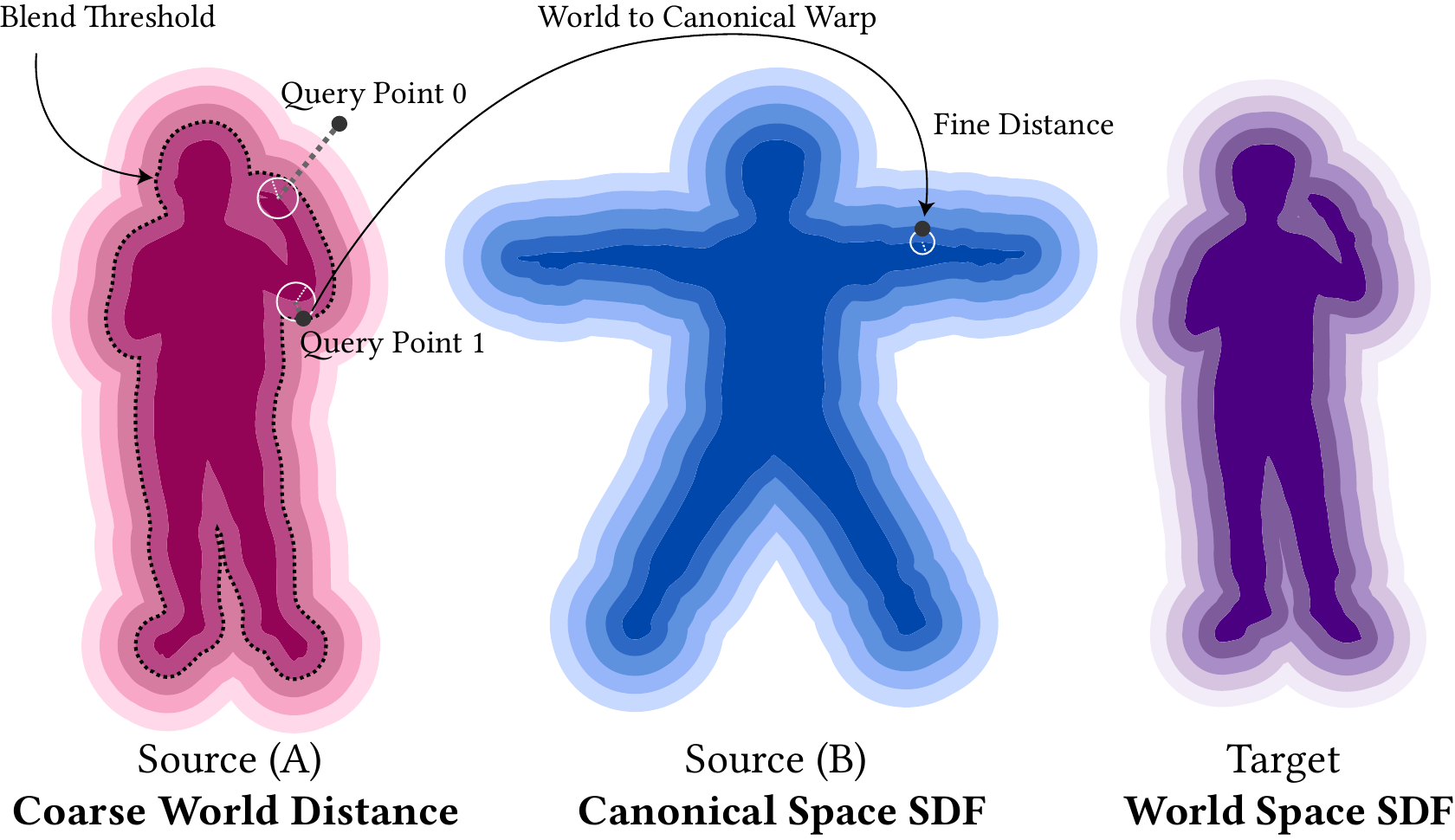}
    \caption{
        \textbf{Illustration of the Hierarchical Distance Query algorithm.}
        In this figure, query point 0 falls out of the cut-off threshold thus its coarse distance is used directly as the world space distance, while query point 1 blends the coarse world space distance and fine canonical distance together since it is within the range of local SDF values.
    }
    \label{fig:hdq}
\end{figure}

%% file: content/05_experiments.tex
\section{Experiments}

\input{figures/comparison.tex}
\input{tables/comparison}

In this section, we conduct qualitative and quantitative experiments to investigate the performance of our relightable neural avatars. All hyperparameters are fixed through out the experiments unless otherwise specified. In Section~\ref{sec:datasets}, we briefly introduce the datasets used for evaluation. Then we make quantitative and qualitative comparisons with three baseline methods in Section~\ref{sec:baseline_comparisons}. Finally, we conduct ablation studies to investigate the effectiveness of our Hierarchical Distance Query and the soft visibility scheme in Section~\ref{sec:ablation_hdq}.

\subsection{Datasets}
\label{sec:datasets}

We collect two datasets \textit{MobileStage} and \textit{SyntheticHuman++} for evaluation. \textit{MobileStage} is a real-world multi-view (36 views) dataset created with synchronized mobile phone cameras on 4 real-world humans. We uniformly select 12 views for training. \textit{SyntheticHuman++} contains 4 sequences (20 views) of dynamic 3D human models with ground truth shape and relighting information. We uniformly select 10 views for training for the sparse-view setting and we use the fourth view for the monocular setting. Please refer to the supplementary material for more detail.

\subsection{Baseline Comparisons}
\label{sec:baseline_comparisons}

\paragraph{Baselines.} To the best of our knowledge, there are very few prior works that study the exact same setting as ours, i.e. training with unknown illumination and sparse-view (or monocular) videos while rendering with novel illumination and novel human poses. We take NeRFactor \cite{zhang2021nerfactor} and Relighting4D \cite{chen2022relighting4d} as baselines and make comparisons with them on both real and synthetic datasets. Since NeRFactor is designed to handle static objects, we only train and evaluate it on the multi-view images of the first frame of each video. We observe that their normal and visibility MLPs often fail under complex human motions, thus we additionally compare with a Relighting4D* and NeRFactor* variant where we use the normal and visibility computed from the density MLP instead of the normal and visibility MLPs.
To illustrate the effectiveness of our proposed components, we additionally render AniSDF \cite{peng2022animatable} as a baseline.

\paragraph{Metrics.} For quantitative analysis, we compare the normal (in degrees), albedo (in PSNR, SSIM and LPIPS \cite{zhang2018unreasonable}), light visibility rendered with uniform BRDF (in PSNR, SSIM, and LPIPS) and relighting (in PSNR, SSIM, and LPIPS) results on 6 different light probes obtained from Polyhaven \cite{polyhaven}.
Following \cite{zhang2021physg}, we align the diffuse albedo and rendered images with ground truth ones before computing metrics 
to mitigate the inherent scale ambiguity in the inverse rendering problem. Note that we compute PSNR using the save protocal as \cite{peng2021animatable}, only computing metrics on the human region. When computing metrics on the full image, our method reports PSNR of 28-30 dB for the relighting and visibility task.
We do not compare the roughness term since Blender uses a different Principled BRDF model from \cite{walter2007microfacet}. Environment map of \textit{SyntheticHuman} \cite{peng2022animatable} is not available since they used programmatically defined light sources. Qualitative results can be found in the supplementary video.
We compare the uniform shading results to evaluate the visibility quality, where the BRDFs of the reconstructed avatars are set to 0.8 across all radiance directions (denoted ``Visibility'') when rendering. Since \textit{SyntheticHuman} \cite{peng2022animatable} does not provide ground truth models for novel poses, we only perform quantitative comparisons on training poses in Table~\ref{tab:comparison}, while qualitative analysis of animating the avatars can be found in Figure~\ref{fig:comparison} and the supplementary video.

\paragraph{Results.} As shown in Figure~\ref{fig:comparison}, our approach can successfully decompose the material and dynamic geometry of the neural avatar, generating a relightable neural avatar from only sparse-view (or monocular) video inputs. In comparison, NeRFactor\cite{zhang2021nerfactor} trained on 1 video frame overfits the training image when training views are sparse.
Relighting4D \cite{chen2022relighting4d} passes structured latent codes \cite{peng2021neural} to NeRFactor's MLPs, enabling it to relight a dynamic video of human performance. However, its quality decreases greatly when synthesizing novel poses. This is mainly because the visibility and normal MLP used in \cite{chen2022relighting4d} is not generalizable to novel human motions. For the Relighting4D* variant, the density backbone still fails to generalize to novel poses \cite{peng2022animatable,liu2021neural}. AniSDF \cite{peng2022animatable} bakes illuminations effects like self-occlusions onto the rendering network, thus the reconstructed neural avatar
looks unrealistic under novel illuminations. Qualitative results on monocular inputs can be found in Figure~\ref{fig:synthetic_human_novel} and the supplementary video.
Relighting4D and NeRFactor take 3s to render a 512$\times$512 image, their ``*'' variants take 50s and our method takes 5s.

\subsection{Ablation Studies}
In this part, we ablate the effectiveness of our proposed Hierarchical Distance Query and soft visibility scheme on relighting quality with the \textit{jody} model of \textit{SyntheticHuman++} under the sparse-view setting. We provide more detailed ablation on the soft visibility scheme, the number of sphere tracing steps, the number of vertices for GS-KNN, and the number of canonical material samples in the supplementary.

\paragraph{Effectiveness of Hierarchical Distance Query and soft visibility scheme.}
\label{sec:ablation_hdq}
\input{tables/ablation_hdq}
\input{figures/ablation_hdq_comparison}

In Figure~\ref{fig:ablation_hdq_comparison}, We compare the results of performing sphere tracing on the canonical space distance (``w/o $d_{coarse}^{world}$''), coarse GS-KNN distance (``w/o $d_{fine}^{can}$'') and our proposed hierarchically queried distance (``Ours'').
As shown in the figure, the canonical space distance is incorrect when the query point is far from the actual surface of the human geometry, resulting in incorrect surface intersection points after the termination of the sphere tracing algorithm. Additionally, computing light visibility on this incorrect distance field would lead to false black regions since distances far from surface points are not reported correctly. Performing surface intersection and visibility computation on the coarse distance results in distorted rendering results. The ``w/o HDQ'' variant uses volume rendering of 128 samples per ray for surface intersection and light visibility computation, leading to an excessive rendering time of 60s per image for a resolution of 512$\times$512, while our HDQ algorithm is able to obtain 10x speed-up at 5s per image with superior rendering quality.

We demonstrate the effectiveness of our HDQ-DFSS algorithm by comparing it with two other variants where (a) hard visibility is used (``w/o soft vis'') and (b) only local visibility computed from normal is used (``w/o HDQ vis''). The quantitative comparison of all three variants can be found in Table~\ref{tab:ablation_hdq}. Note that although the ``w/o soft'' variants report higher PSNR and SSIMs, the visual quality of hard cast shadows is worse than ours, as indicated by the LPIPS metric and visible in Figure~\ref{fig:ablation_hdq_comparison}.

\paragraph{Sensitivity analysis on hyper-parameters.} 
We provide additional sensitivity analysis of hyper parameters including $N_{st}$ (the number of sphere tracing steps), $N_{st}^{vis}$ (the number of sphere tracing steps for visibility computation), $K$ (the number of points sampled to perform GS-KNN and canonical-to-world warping), $N_{s}$ (the number of points to sample in the material MLPs) and a runtime and quality analysis regarding the cut-off value $\tilde{T}_d$ in the supplementary.

%% file: figures/comparison.tex
\begin{figure*}[ht]
    \centering
    \includegraphics[width=1.0\textwidth]{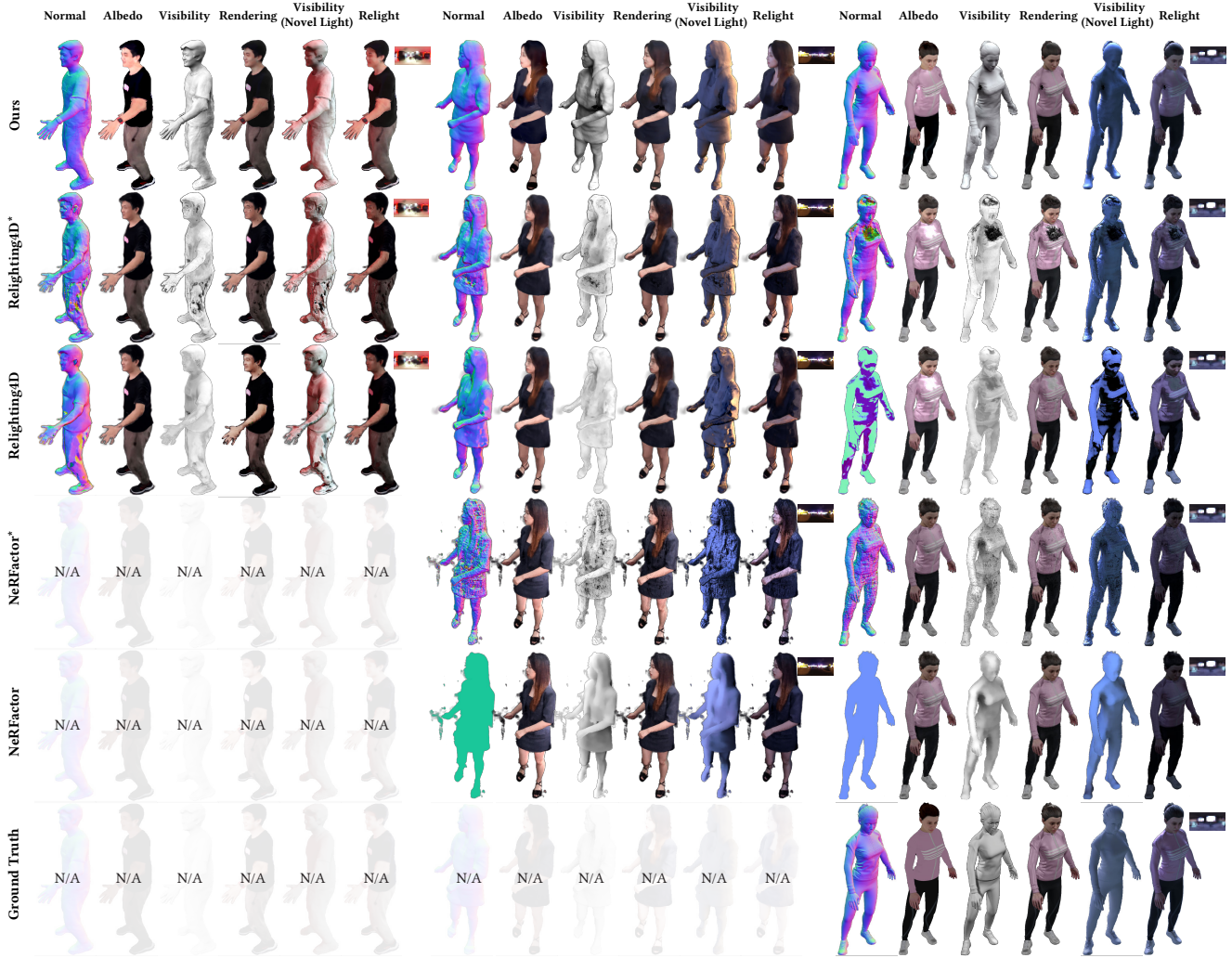}
    \caption{\textbf{Qualitative comparison of our method and baselines.} 
    The first six columns display the results of synthesizing a character in a novel pose from the \textit{MobileStage} dataset. The middle six columns depict a character in a training pose from the \textit{MobileStage} dataset. For the last six columns, we show results from \textit{SyntheticHuman++}, for which we have ground truth as reference. Note that NeRFactor is only trained on 1 frame. Relighting4D* and NeRFactor* denote directly computing normal and visibility using their density MLPs.} 
    \label{fig:comparison}
\end{figure*}

%% file: tables/comparison.tex
\begin{table*}[ht]
    \caption{
        \textbf{Quantitative comparison.} We compare our method with baselines on the \textit{SyntheticHuman++} dataset.
        Following \cite{peng2021neural}, the SSIM and LPIPS \cite{zhang2018unreasonable} metrics are computed in the bounding box of the human region, while the degree of normal and PSNR metrics are computed within the foreground mask.
        Due to the inherent scale ambiguity in the inverse rendering task, we align the rendered images and albedo with ground truth ones following \cite{zhang2021physg} before computing metrics.
        Since NeRFactor \cite{zhang2021nerfactor} is designed to fit static objects, we train and evaluate it only on the first frame of each sequence.
        All image metrics are computed in the foreground region in linear color space.
        ``$*$'' denotes variants without the normal and visibility MLPs.
        AniSDF produces baked lighting in the rendering network, while NeRFactor \cite{zhang2021nerfactor} and Relighting4D \cite{chen2022relighting4d} perform poorly on the uniform shading task for visibility since the reconstructed geometry is too rough. Our method works well even on the challenging monocular setting.
    }
    \label{tab:comparison}
    \centering

    \addtolength{\tabcolsep}{-2pt}
    \resizebox{0.95\linewidth}{!}{

        \begin{tabular}{llcccccccccc} %
            \toprule

                                                  &                       & \multicolumn{1}{c}{\textbf{Normal}} & \multicolumn{3}{c}{\textbf{Diffuse Albedo}} & \multicolumn{3}{c}{\textbf{Relighting}} & \multicolumn{3}{c}{\textbf{Visibility}}                                                                                                             \\
                                                  &                       & Degree $\downarrow$                 & PSNR$\uparrow$                              & SSIM$\uparrow$                          & LPIPS$\downarrow$                       & PSNR$\uparrow$ & SSIM$\uparrow$ & LPIPS$\downarrow$ & PSNR$\uparrow$ & SSIM$\uparrow$ & LPIPS$\downarrow$ \\
            \midrule
            \multirow{6}{*}{\textbf{Sparse-View}} & Ours                  & \textbf{12.44}                      & \textbf{29.01}                              & \textbf{0.933}                          & \textbf{0.119}                          & \textbf{22.69} & \textbf{0.861} & \textbf{0.206}    & \textbf{20.20} & \textbf{0.848} & \textbf{0.155}    \\
                                                  & Relighting4D*         & 29.38                               & 24.70                                       & 0.885                                   & 0.183                                   & 22.13          & 0.835          & 0.237             & 15.22          & 0.763          & 0.252             \\
                                                  & Relighting4D          & 93.83                               & 24.71                                       & 0.885                                   & 0.183                                   & 20.87          & 0.774          & 0.276             & 5.366          & 0.514          & 0.375             \\
                                                  & NeRFactor*  (1 frame) & 34.29                               & 22.23                                       & 0.817                                   & 0.226                                   & 21.04          & 0.758          & 0.313             & 11.37          & 0.581          & 0.387             \\
                                                  & NeRFactor  (1 frame)  & 51.92                               & 22.23                                       & 0.817                                   & 0.226                                   & 20.70          & 0.757          & 0.299             & 10.56          & 0.597          & 0.361             \\
                                                  & AniSDF                & 14.72                               & 22.13                                       & 0.862                                   & 0.202                                   & 17.55          & 0.799          & 0.262             & -              & -              & -                 \\
            \midrule
            \multirow{4}{*}{\textbf{Monocular}}   & Ours                  & \textbf{18.71}                      & 23.42                                       & \textbf{0.873}                          & \textbf{0.176}                          & \textbf{22.45} & \textbf{0.831} & \textbf{0.224}    & \textbf{17.95} & \textbf{0.761} & \textbf{0.212}    \\
                                                  & Relighting4D*         & 26.17                               & \textbf{25.37}                              & 0.864                                   & 0.210                                   & 21.81          & 0.802          & 0.254             & 17.10          & 0.709          & 0.286             \\
                                                  & Relighting4D          & 81.74                               & 25.36                                       & 0.864                                   & 0.210                                   & 21.85          & 0.806          & 0.268             & 16.18          & 0.726          & 0.302             \\
                                                  & AniSDF                & 20.36                               & 21.51                                       & 0.812                                   & 0.255                                   & 18.29          & 0.745          & 0.297             & -              & -              & -                 \\
            \bottomrule
        \end{tabular}
    }
    \addtolength{\tabcolsep}{2pt}

\end{table*}

%% file: tables/ablation_hdq.tex
\begin{table}[ht]
    \caption{
        \textbf{Ablation study on the proposed Hierarchical Distance Query algorithm.}
        Qualitative results can be found in Figure~\ref{fig:ablation_hdq_comparison}.
        The ``w/o HDQ'' variant uses naive volume rendering (128 samples per ray) to compute pixel-surface intersection and visibility, which is not only slow (60s per image at 512 $\times$ 512 resolution compared to our 5s per image) but also exhibits inferior rendering quality.
    }
    \label{tab:ablation_hdq}
    \centering
    \addtolength{\tabcolsep}{-5pt}

    \resizebox{1.0\linewidth}{!}{
        \begin{tabular}{lcccccc}
            \toprule
            \multicolumn{1}{c}{}          & \multicolumn{3}{c}{\textbf{Relighting}} & \multicolumn{3}{c}{\textbf{Visibility}}                                                                           \\
                                          & PSNR$\uparrow$                          & SSIM$\uparrow$                          & LPIPS$\downarrow$ & PSNR$\uparrow$ & SSIM$\uparrow$ & LPIPS$\downarrow$ \\
            \midrule
            Ours                          & \textbf{21.57}                          & \textbf{0.853}                          & \textbf{0.168}    & 20.53          & 0.869          & \textbf{0.142}    \\
            w/o soft vis                  & 21.19                                   & 0.848                                   & 0.173             & \textbf{21.27} & \textbf{0.873} & 0.145             \\
            w/o HDQ vis                   & 21.00                                   & 0.844                                   & 0.175             & 20.88          & 0.869          & 0.143             \\
            w/o HDQ                       & 21.36                                   & 0.753                                   & 0.196             & 20.00          & 0.760          & 0.173             \\
            w/o $d_{coarse}^{world}$      & 20.69                                   & 0.792                                   & 0.236             & 18.75          & 0.767          & 0.250             \\
            w/o $d_{fine}^{can}$ (GS-KNN) & 19.56                                   & 0.784                                   & 0.245             & 14.63          & 0.758          & 0.233             \\
            \bottomrule
        \end{tabular}
    }
    \addtolength{\tabcolsep}{5pt}

\end{table}

%% file: figures/ablation_hdq_comparison.tex
\begin{figure*}[ht]
    \centering
    \includegraphics[width=0.985\textwidth]{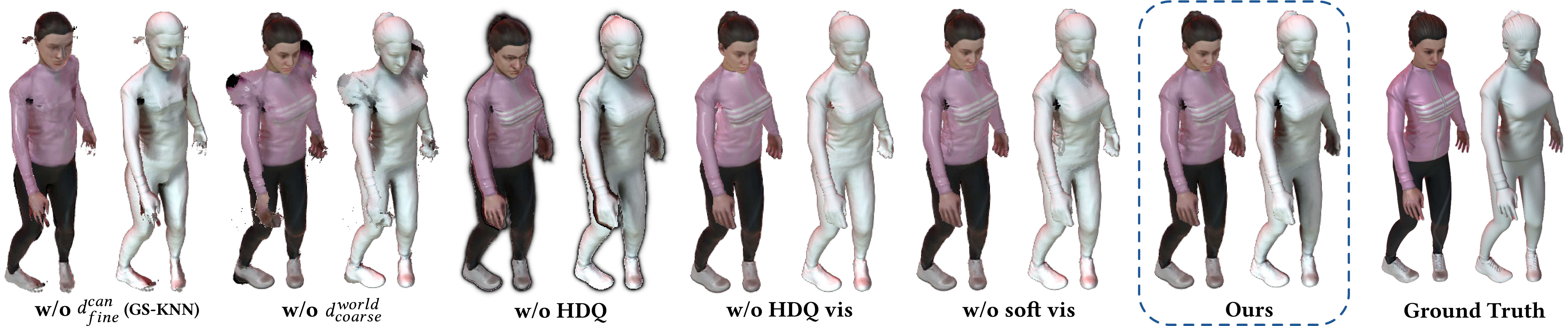}
    \caption{
        \textbf{Effectiveness of Hierarchical Distance Query}. Performing sphere tracing using only the canonical distance $d_{fine}^{can}$ or coarse world distance $d_{coarse}^{world}$ results in incorrect surface intersection $\boldsymbol{x}_s$ and soft visibility $p_s$, while tracing with our proposed Hierarchical Distance Query produces correct results. Using hard shadow (``w/o soft vis'') or local shadow (``w/o HDQ vis'') leads degraded perceptual quality.
    }
    \label{fig:ablation_hdq_comparison}
\end{figure*}

%% file: content/06_conclusion.tex
\section{Conclusion and Discussion}
This paper presents a novel framework to reconstruct relightable and animatable neural avatars from only sparse-view (or monocular) video input. We generalize the canonical distance field to arbitrary human poses via a hierarchical distance query scheme, with which the photometric properties of the neural avatar can be easily retrieved for relighting. We demonstrate that together with other innovative components, our approach reconstructs high-quality animatable geometry and material, supporting realistic relighting.
This work also has some limitations. Although the proposed method produces high-quality relighting results from challenging sparse-view or monocular settings, it has the natural limitation of neural field methods in that it requires a long training time of 20 hours.
Future work could consider recent neural field acceleration methods to further increase the training speed.

%% file: content/09_supplementary.tex
\section{Addtional Results}

\subsection{Addtional Comparisons}
In this section, we provide more qualitative results. Dynamic reconstruction and animation results under novel illuminations can be found in the supplementary video, which better demonstrates the effectiveness of our method to reconstruct the animatable human avatar than figures.

We provide more relighting and animation results in Figure~\ref{fig:synthetic_human_novel} and Figure~\ref{fig:mobile_stage_novel} where we show the reconstructed neural avatar being driven by novel human poses extracted from the AIST++ \cite{li2021ai} dataset and relight with HDRi images from Polyhaven \cite{polyhaven}.

Additionally, we provide comparisons on monocular inputs for our method and baselines in Figure~\ref{fig:comparison_mono}. Relighting4D \cite{chen2022relighting4d} failed to reconstruct the normal and visibility MLP under this challenging monocular input, leading to incorrect relighting results and visibility estimation. Our method is able to recover the relightable properties and successfully relight the human avatar given even only monocular input.
\input{figures/comparison_mono}

\subsection{Additional Ablation Studies}

\paragraph{Effectiveness of soft visibility scheme.}
\input{figures/ablation_hdq_distance}
\label{sec:ablation_visibility}
\input{tables/ablation_dfss}
\input{figures/ablation_dfss}

We also study the effect of introducing the soft visibility scheme in Figure~\ref{fig:ablation_dfss} and Table~\ref{tab:ablation_dfss}. All hyperparameters are fixed through out the experiments unless otherwise specified.
We show the results of lighting a model with uniform material, where the BRDFs are equal everywhere, using a single area light source in Figure~\ref{fig:ablation_dfss}. This effectively produces ambient occlusion for the models. The quantitative comparison of all three variants can be found in Table~2 of the main paper.
Note that although the ``w/o soft'' variants report higher PSNR and SSIMs, the visual quality of hard cast shadows is worse than ours, as indicated by the LPIPS metric.
Additionally, we compare the effect of modeling visibility on material acquisition in Figure~\ref{fig:ablation_vis_albedo}. The results indicate not modeling visibility leads to baked shadow on the diffuse albedo, while our method reconstructs a much more uniform diffuse map.

\paragraph{Sensitivity analysis and ablation study on other hyperparameter choices.}

In this section, we ablate the influence of other hyperparameters used in the model. The quantitative results are shown in Table~\ref{tab:ablation_dfss}. $N_{st}$ denotes the number of sphere tracing steps for finding pixel-surface intersections. Table~\ref{tab:ablation_dfss} shows that too low a number of steps for sphere tracing would lead to worse rendering quality, however, a too high number of steps only leads to diminishing returns. Thus we chose $N_{st} = 16$ in all our experiments. $N_{st}^{vis}$ is the number of sphere tracing steps for computing the soft visibility term, which shares a similar diminishing return pattern as $N_{st}$. The number of vertices $K$ used in the inverse warping process and the number of samples $N_{s}$ taken in canonical space for material properties (Section~\ref{sec:material_volume}) does not significantly impact the rendering quality. Thus their values are empirically set to $10$ and $3$ in our experiments.

Furthermore, we provide a runtime and metric analysis on different choices of cut-off value $\tilde{T}_d$ used during the sphere tracing of the neural distance field in Table~\ref{tab:ablatin_cutoff}. In our method, the cut-off during camera-human intersection sphere-tracing ($\tilde{T}_d$) is set to $0.1$ and the cut-off for computing the visibility term ($\tilde{T}_d^{vis}$) is set to $0.025$. It's worth noting that setting the cut-off value to zero is effectively only using the coarse distance produced by GS-KNN for rendering, which greatly degrades the quality of relighting and rendering as shown in Figure~5 of the main paper.

\paragraph{Accuracy of the proposed Hierarchical Distance Query and sphere tracing algorithm.}
\input{tables/ablation_sdf}

\input{figures/hdq_geo}

\input{tables/ablation_cutoff}

In Table~\ref{tab:ablation_sdf}, we provide an additional ablation study on the accuracy of the proposed HDQ-ST (Hierarchical Distance Query and sphere tracing) algorithm. ``w/o HDQ'' variant uses volume rendering of 128 samples per ray for surface intersection and ``w/o $d_{fine}^{can}$'' and ``w/o $d_{coarse}^{world}$'' denote variant where sphere tracing is performed without the fine distance and coarse distance of the HDQ algorithm respectively. Our method achieves at least two orders of magnitude higher accuracy at $0.00017$m compared to other variants.

\section{Additional Implementation Details}

\subsection{Details of the Datasets}

\paragraph{MobileStage.} To evaluate our approach on real-world humans, we collect a real-world multi-view dataset called \textit{MobileStage}. This dataset is composed of 4 dynamic human videos collected using 36 synchronized mobile phone cameras at 1920$\times$1080 resolution and 30 fps with complex indoor environment illuminations.
We uniformly select 12 views for training in all experiments, the detailed frames selected are listed in Table~\ref{tab:training_settings}.
The characters perform complex motions, including talking, walking, and swinging, resulting in complex shading and shadow effects.
We calibrate the camera poses of the multi-view system using COLMAP \cite{schonberger2016structure} and extract foreground human mask with \cite{lin2022robust}. The human poses corresponding to each of the characters are estimated using EasyMocap \cite{easymocap}.

\paragraph{SyntheticHuman++}
\textit{SyntheticHuman++} contains 5 sequences of dynamic 3D human models with detailed shapes and complex materials. We use the original multi-view video in SyntheticHuman \cite{peng2022animatable} for training, where 10 views of the synthetic characters were rendered with obvious self-occlusion and shading variations using Blender Cycles \cite{blender2018}. We uniformly select 6 views from all 20 views for testing. For the monocular setting, we select the fourth view for training.

To measure the quality of the visibility term, we additionally render all sequences with uniform diffuse material where the BRDFs in all directions are the same. This factors out the material from the rendering algorithm.
For evaluating the performance of inverse rendering photometric properties, we also render the ground truth albedo and normal.
We do not compare with the roughness term since Blender uses a different material model than ours.
For novel pose synthesis, we used the motion capture data from AIST++ \cite{li2021ai} along with motion capture data from the ZJU-MoCap Dataset \cite{peng2021neural, fang2021mirrored} (sequence \textit{377}, \textit{386}, \textit{390}).

\input{tables/dataset_setting}
\input{figures/ablation_vis_albedo}

\subsection{Implementation Details for GS-KNN}

Since vanilla KNN only produces unsigned distance to the closest $K$ points, the sign of the distance value is determined by the dot product between $\boldsymbol{x} - v_k$ and the world space normal $\boldsymbol{n}_k$ corresponding to $i_k$ using
\begin{equation}
    d_k = \mathrm{sign}(\mathrm{dot}(\boldsymbol{x} - \boldsymbol{v}_k, \boldsymbol{n}_k))  d_k.
\end{equation}
Since the world space query point might be close to multiple parts of the posed vertices of the world space parametric model, which will lead to incorrect inverse LBS results, we need to avoid using KNN query results from vertices that are far away in the canonical space.
To avoid assigning the query point to multiple parts of the posed vertices $\mathcal{V}$ of the world space parametric model, we further update the KNN results using an approximation of the geodesic distance on the parametric human model.
Specifically, we take the index of the closest parametric model vertex $i_{min}$, and compute the approximated canonical Euclidian distance between all other $K-1$ selected vertices by querying their distance $d^{\mathcal{V}}_{k}$ on the canonical parametric model $\mathcal{V}^{can}$ (e.g. T-Pose) using
\begin{multline}
    i_k, d_k, \boldsymbol{v}_k, \boldsymbol{n}_k, \boldsymbol{w}_k = \\
    \begin{cases}
        i_{min}, d_{min}, \boldsymbol{v}_{min}, \boldsymbol{n}_{min}, \boldsymbol{w}_{min} & , \text{if } d^{\mathcal{V}}_{k} < T_d. \\
        i_k, d_k, \boldsymbol{v}_k, \boldsymbol{n}_k, \boldsymbol{w}_k                     & , \text{otherwise}.
    \end{cases}
\end{multline}
where $d_{min}$, $\boldsymbol{v}_{min}$, $\boldsymbol{n}_{min}$, $\boldsymbol{w}_{min}$ are the vertex properties corresponding to $i_{min}$, and $T_{d}$ is the geodesic distance cutoff threshold, which is set to $0.1$ through all experiments.

\subsection{Implementation Details for Inverse LBS}
With $K$ sets of vertex properties produced by GS-KNN, the world-space query point is transformed into canonical space by the inverse LBS module. The human pose defines $B$ body parts, which produces $B$ transformation matrices $\{{G}_b\} \in SE(3)$. In the inverse LBS module, the world space points are transformed to canonical space using
\begin{equation}
    \label{eq:inverse_lbs_transform}
    \boldsymbol{T}^{world}(\boldsymbol{x}) = (\sum_{b=1}^{B}w_b(\boldsymbol{x}){G}_b)^{-1},
\end{equation}
\begin{equation}
    \label{eq:inverse_lbs}
    \boldsymbol{x^{\prime}} = \boldsymbol{T}^{world}(\boldsymbol{x})\boldsymbol{x},
\end{equation}
where the blend weights $w_b$ of the $b$th body part is the $b$th element of the blend weight vector $\boldsymbol{w} = (w_1,...,w_B)$, which is computed from the KNN results $\mathcal{D}_K=\{d_0...,d_k\}$ and $\mathcal{W}_K=\{\boldsymbol{w}_0...,\boldsymbol{w}_k\}$ using
$\mathcal{D}_K=\{d_0...,d_k\}$ and $\mathcal{W}_K=\{\boldsymbol{w}_0...,\boldsymbol{w}_k\}$:
\begin{equation}
    \label{eq:blend_weights_blending}
    \boldsymbol{w} = \sum_{k=1}^{K} \mathrm{softmax} (\frac{-d_0} {2  R_w^2}, ..., \frac{-d_K} {2  R_w^2})_{k}  \boldsymbol{w}_k,
\end{equation}
where $\mathrm{softmax}$ denotes the softmax normalization operator, and $R_w$ is the blend weight blending radius, which is set to $0.075$ through all experiments following the distance-weighted blending \cite{zheng2022structured}.

\begin{equation}
    \label{eq:canonical_space}
    \boldsymbol{x} = \boldsymbol{x}^\prime + F_{\Delta{\boldsymbol{x}}}(\Theta_f, \boldsymbol{x}^\prime).
\end{equation}

Using Equation~\ref{eq:inverse_lbs} and Equation~\ref{eq:blend_weights_blending}, the world space query point $\boldsymbol{x}$ is transformed to canonical space $\boldsymbol{x^{\prime}}$.

For world space normal direction, we compute the normalized gradient of the canonical deformed SDF and transform it back to world space using:
\begin{equation}
    \label{eq:backward_normal}
    \boldsymbol{n}_{s}(\boldsymbol{x}_{s}) = (\boldsymbol{R}^{world})^{-1} (\nabla S(\boldsymbol{x}_{s}^{can})),
\end{equation}
where $\boldsymbol{R}^{world}$ is the rotation component of $\boldsymbol{T}^{world}$ the inverse warping step of HDQ, and $\boldsymbol{x}_{s}^{can} = \boldsymbol{T}^{world}\boldsymbol{x}_{s}$ is the corresponding canonical space coordinate of the surface intersection $\boldsymbol{x}_{s}$.

\subsection{Listing of the Surface Intersection Sphere Tracing Algorithm}
In Algorithm~\ref{alg:surface_intersection_sphere_tracing} we provide the detailed procedure for finding surface intersections when performing sphere tracing on the hierarchically queried distance.
The sphere tracing procedure on ray $\boldsymbol{r}(t) = \boldsymbol{o} + t \boldsymbol{d}$ is performed with $t$ bounded by near and far distances $n$ and $f$. Additionally, we add an offset $o$ to the queried distance when updating $t$ to skip through gazing angles, and we linearly interpolate for the surface intersection $\boldsymbol{x}_s$ when the sign of the queried distance changes. 
We empirically set $N_{st} = 16$ and $o = 0.02$ for all experiments. $n$ and $f$ are provided by intersecting the camera ray with the axis-aligned bounding box of the posed parametric human model corresponding to the neural avatar.

\input{algos/geo_st}

\subsection{Listing of the Sphere Tracing Soft Visibility Algorithm}
Light visibility is defined as whether a light ray from the light source to the surface intersection point is occluded by other parts of the geometry.
This occlusion test can be achieved by performing the same Sphere Tracing algorithm for surface intersection and checking whether the returned intersection point $\boldsymbol{x}_s$ lies on the far plane $f$.

However, such a simple binary test can only produce hard shadow for a point light source \cite{chen2022tracing}, while perfect point light rarely exists and the more common area lights should produce soft shadows \cite{zhang2021nerfactor}. If we simply perform dense integration or monte-carlo integration to compute the soft light visibility values, the computational cost will be too high. To this end, we propose to integrate traditional distance field soft shadow algorithms \cite{parker1998single,ban2019area,aaltonen2018gpu} to our hierarchically queried driable neural distance field, which interpret the global meaning of the distance values as the penumbra coefficient for soft shadow effects with a small number of $\boldsymbol{\omega}_i$ samples.

\input{algos/dfss_st}

Algorithm~\ref{alg:soft_light_visibility_sphere_tracing} illustrates the soft visibility computation algorithm, in which $\boldsymbol{o}$ is set to the surface intersection $\boldsymbol{x}_s$, $\boldsymbol{d}$ is set to the light direction $\boldsymbol{\omega}_i$, $n$ is set to $0.01$, $f$ is set to $10.0$ and $N^{vis}_{st}$ is set to 4 in all experiments.
The ratio between the current distance $d_1$ and current ray depth $t$ along with the area $a$ (in solid angle) of the light source forms $p_{s} = \frac{d_1}{2 t \sqrt{\frac{a}{\pi}}}$ following \cite{parker1998single,ban2019area}.

\subsection{Details on Network Queries of the Canonical Material Fields}
\label{sec:material_volume}
To apply more supervision on the rendering process and the canonical material MLPs, we construct a sparse set of volume sampling $\mathcal{T}_{s}$ along the camera ray $\boldsymbol{r}(t) = \boldsymbol{o} + t \boldsymbol{d}$ near the computed surface intersection point $\boldsymbol{x}_s = \boldsymbol{o} + t_s \boldsymbol{d}$ with a fixed step size $t_{step} = 0.005$ and number of samples $N_s = 3$ using
\begin{equation}
    \mathcal{T}_{s} = t_s + \{ 2\frac{n_s}{N_s} t_{step},...,2\frac{N_s}{N_s} t_{step} \} - t_{step}.
\end{equation}
Then, $\mathcal{T}_{s}$ is used to sample the material networks, on which the Volume Rendering algorithm in \cite{peng2022animatable,yariv2021volume} is applied to compute the final albedo $\alpha_{s}(\boldsymbol{x}_s)$ and roughness $\gamma_s(\boldsymbol{x}_s)$ corresponding to $\boldsymbol{x}_s$.

\subsection{Network Structures}

\input{figures/network_structure}

For the canonical space geometry and displacement field, we use 8 layer MLPs of width 256 for $S$ and $F_{\delta\boldsymbol{x}}$ with ReLU and Softplus activation respectively. $S$ takes positionally encoded (PE) \cite{mildenhall2021nerf} coordinate of resolution 8 as input and $F_{\delta\boldsymbol{x}}$ takes point input with 10 levels of PE along with the pose of the current human frame. We follow \cite{peng2022animatable} to initialize $S$ with \cite{gropp2020implicit}.
For canonical material networks, we use 8-layer MLPs for $A$ and $\Gamma$ with ReLU activation of points input with 10 levels of PE. We additionally input the current human pose to increase its representational power.
An illustration of the network structure is shown in Figure~\ref{fig:network_structure}.

\subsection{Loss Functions}
We optimize our relightable and animatable neural human avatar by rendering the image with given camera poses and compare the pixel values $L_o$, which corresponds to ray $\boldsymbol{r}$, where $ \boldsymbol{x}_s =\boldsymbol{o} + t_s \boldsymbol{d}$ and $\boldsymbol{\omega}_o = \boldsymbol{d}$, against the ground truth ones $L_{gt}$. The loss function is defined as
\begin{equation}
    \mathcal{L}_{d} =\sum_{\boldsymbol{r} \in \mathcal{R}}  {\lVert L_o(\boldsymbol{r})- L_{gt}(\boldsymbol{r}) \rVert}_2.
\end{equation}
where $\mathcal{R}$ denotes all camera rays in the forward rendering process.

Following \cite{peng2022animatable}, we additionally regularize the residual displacement field $F_\Delta$ and canonical SDF $S$ with the Eikonal term using
\begin{multline}
    \mathcal{L}_{eik} = \\
    \sum_{\boldsymbol{x} \in \mathcal{X}} \lambda_{e0}( {\lVert \Delta_{\boldsymbol{x}} S(\boldsymbol{x} + \Delta F(\boldsymbol{x})) \rVert}_2 - 1) + \\
    \lambda_{e1} ({\lVert \Delta_{\boldsymbol{x} + \Delta F(\boldsymbol{x})} S(\boldsymbol{x} + \Delta F(\boldsymbol{x})) \rVert}_2 - 1)
\end{multline}
where $\mathcal{X}$ denotes all the warped canonical points meeting the cutoff criteria $d_{coarse}^{world} <= \tilde{T}_d$ during the forward rendering process.

Their \cite{peng2022animatable} regularization on the magnitude of the residual displacement field $F_\Delta$ is also adopted in our work, which is defined as
\begin{equation}
    \mathcal{L}_{r} = \sum_{\boldsymbol{x} \in \mathcal{X}}  {\lVert F_\Delta(\boldsymbol{x}) \rVert}_2.
\end{equation}

We also define a mean intersection of union foreground mask loss on the silhouette $M_{o}(\boldsymbol{r})$ of the rendered image to regularize the training process as
\begin{equation}
    \mathcal{L}_{m} = \sum_{\boldsymbol{r} \in \mathcal{R}}  \frac { M_{o}(\boldsymbol{r})  M_{gt}(\boldsymbol{r})} {M_{o}(\boldsymbol{r}) + M_{gt}(\boldsymbol{r})}.
\end{equation}

To alleviate ambiguities in the physically-based rendering model, we follow \cite{chen2022relighting4d} to add sparsity and smoothness regularizations on our material representation, the albedo MLP $\alpha = A(\boldsymbol{x})$ and the roughness MLP $\gamma = \Gamma(\boldsymbol{x})$. The regularization terms are defined as
\begin{equation}
    \mathcal{L}_{ent} = \mathrm{gaussian\_entropy}(\alpha(\mathcal{X}))
\end{equation}
\begin{equation}
    \mathcal{L}_{a} = \sum_{\boldsymbol{x} \in \mathcal{X}}  {\lVert  \alpha(\boldsymbol{x}) - \alpha(\boldsymbol{x} + \Delta\boldsymbol{x}) \rVert}_2
\end{equation}
\begin{equation}
    \mathcal{L}_{r} = \sum_{\boldsymbol{x} \in \mathcal{X}}  {\lVert  \gamma(\boldsymbol{x}) - \gamma(\boldsymbol{x} + \Delta\boldsymbol{x}) \rVert}_2
\end{equation}
where $\mathrm{gaussian\_entropy}(\alpha(\mathcal{X}))$ is the entropy of the Gaussian distribution of the albedo map $\alpha$ evaluated on the canonical points $\mathcal{X}$ as in \cite{chen2022relighting4d} and $\Delta \boldsymbol{x}$ is a random perturbation sampled from a normal distribution $\Delta \boldsymbol{x} \sim \mathcal{N}(\mu,\sigma^{2})$ with $\mu = 0$ and $\sigma = 0.02$ in all experiments.

\subsection{Training}
To make the optimization process more controllable, we separate the training process into two stages by postponing the training of the material and environment light probe after the geometry of the neural avatar has converged.

We first train the base geometry of the neural avatar including $S$, $\Delta F$ and a neural rendering network $C(\boldsymbol{x},\boldsymbol{d}^{\prime})$ as in \cite{peng2022animatable} with the same volume rendering algorithm defined in \cite{peng2022animatable}.
The rendering network $C(\boldsymbol{x},\boldsymbol{d})$ \cite{peng2022animatable} takes the canonical query point $\boldsymbol{x}$ obtained in Equation~\ref{eq:canonical_space} and the canonical view direction $\boldsymbol{d}^{\prime} = \boldsymbol{R}^{world} \boldsymbol{d}$ as input where $\boldsymbol{R}^{world}$ is the rotation component of the world space transformation matrix $\boldsymbol{T}^{world}$ in Equation~\ref{eq:inverse_lbs_transform}.
The networks are trained using $\mathcal{L}_{d}$, $\mathcal{L}_{eik}$ and $\mathcal{L}_{r}$ for 200k iterations with and Adam optimizer \cite{kingma2014adam} of learning rate $5e^{-4}$ exponentially annealed to $5e^{-6}$ during the first stage training.
The full loss function of the first stage training is defined as
\begin{equation}
    \label{eq:first_stage_loss}
    \mathcal{L}_{1st} = \lambda_{d} \mathcal{L}_{d} + \lambda_{eik} \mathcal{L}_{eik} + \lambda_{r} \mathcal{L}_{r} + \lambda_{m} \mathcal{L}_{m}
\end{equation}
where $\lambda_{d}$, $\lambda_{eik}$, $\lambda_{r}$, $\lambda_{m}$ are set to $1.0$, $0.025$, $0.1$ and $0.01$ respectively. And the loss weights in the eikonal loss $\mathcal{L}_{eik}$ definition $\lambda_{e0}$ and $\lambda_{e1}$ are set to $1.0$ and $2.0$ respectively.

During the second stage of the training process, we replace the volume rendering algorithm of \cite{peng2022animatable} with Sphere Tracing \cite{hart1996sphere} defined on the neural SDF of the human avatar, and the rendering network $C$ is replaced with the physically-based renderer. Moreover, we replace the SMPL model from the first stage with the learned coarse mesh during the second-stage optimization.
Additional losses $\mathcal{L}_{ent}$, $\mathcal{L}_{a}$, $\mathcal{L}_{r}$ are added for regularization with the full loss defined as
\begin{equation}
    \label{eq:second_stage_loss}
    \mathcal{L}_{2nd} = \mathcal{L}_{1st} + \lambda_{ent} \mathcal{L}_{ent} + \lambda_{a} \mathcal{L}_{a} + \lambda_{r} \mathcal{L}_{r}.
\end{equation}
The second stage optimization is performed with an Adam optimizer \cite{kingma2014adam} of learning rate $5e^{-4}$ exponentially annealed to $5e^{-6}$ for 25k iterations. Additionally, the starting learning rate on parameters of $S$ and $\Delta F$ are tuned down to $1e-5$ if not otherwise specified to preserve the geometry during the first few iterations of the second stage. Loss weights $\lambda_{d}$, $\lambda_{eik}$, $\lambda_{r}$, $\lambda_{m}$, $\lambda_{ent}$, $\lambda_{a}$ and $\lambda_{r}$ are set to $10.0$, $0.1$, $0.1$, $0.01$, $5e^{-4}$, $5e^{-3}$ and $5e^{-3}$ respectively, where $\lambda_{e0}$ and $\lambda_{e1}$ are all set to $1.0$.

The first and second stages are all trained with a batch size of 8 with 1024 rays per batch. On two NVIDIA RTX 3090 GPUs, the first stage training (200k iterations) takes about 20 hours and the second stage (25k iterations) takes about 10 hours. The forward rendering process of the second stage is adopted for inference when the neural avatar is driven with novel human poses or relit with novel lighting conditions through all experiments.

\section{Additional Discussion}

\subsection{Discussion on Alternatives to HDQ}
NLST \cite{seyb2019non} also proposes a method on performing sphere tracing on deformed signed distance fields with undeformed-space distance query and integration of an ODE solver. However, their method is not directly applicable to animatable neural avatars since NLST requires the computation of the Jacobian of the inverse deformation field, which is unavailable in the neural deformation field used in our method. One could try to replace the single-direction displacement field with a bi-directional one like in \cite{cai2022neural}, which might reduce the generalizability to novel human poses. But computing the Jacobian of a neural deformation field requires taking the backward of a large computation graph multiple times, which is time-consuming. Combined with the fact that NLST requires multiple subsets in one single query step, it would not be practical to naively combine the two. In comparison, our method does not require computing the Jacobian and directly returns a single well-approximated distance value for an articulated and neurally deformed distance field thanks to the adoption of the observation space Eikonal loss, which essentially also makes the neurally deformed field a valid distance field.

\cite{sharp2022spelunking} could also be considered a viable approach for surface intersection and distance queries if the target of the query is a static neural implicit field. However, its core technique Range Analysis not directly applicable to the animated geometry of our reconstructed neural avatar, which requires computing the closest K points on a parametric model and performing space warping through blended skinning methods.

\subsection{Difference from AniSDF}
Similar to \cite{peng2022animatable}, we define the animatable geometry of the human avatar as the combination of a pose-driven deformation field $F(\boldsymbol{x})$ and a canonical SDF network $S(\boldsymbol{x})$, where the deformation field can be further decomposed into a pose-dependent displacement field $F_{\Delta{\boldsymbol{x}}}$ and KNN-based inverse Linear Blend Skinning (LBS) \cite{lewis2000pose} module.
We follow their canonical and deformation field setup to model the animatable human geometry. However, one key difference is that we focus on producing the correct world-space distance values, instead of simply warping query points back to canonical spaces.
We intend to use the SDF values in a fixed step Sphere Tracing \cite{hart1993sphere} framework where correct world space distance values are expected for convergence.

The key observation that makes the Hierarchical Distance Query scheme possible is that SDF values are locally deformation invariant under the inverse Linear Blend Skinning (LBS) \cite{lewis2000pose} algorithm. The closer the query point is to the geometry, the less the value changes when transforming under novel human poses, where the distance value of a surface point will always be zero no matter the pose.
This makes it possible to use the canonical SDF network $S(\boldsymbol{x})$ for near-surface distance queries, where the accuracy and detail of the distance are well-preserved, and then use the coarse KNN distance values for far-surface distance queries.

Specifically, due to the linearity of the inverse LBS algorithm, close-to-surface points are rarely mapped to the wrong canonical location since they exhibit very little ambiguity, while it is quite likely for far-from-surface points to fall into the region of confusion during the inverse wrapping process. Thus $d_{fine}^{can}$ is rarely inaccurate for close-to-surface query points, while it is more likely to be inaccurate for far-from-surface ones.
Utilizing this property of the SDF value, we can interpret the fine canonical distance value $d_{fine}^{can}$ as a good approximation for the actual world-space fine level distance $d^{world}$ when the query points are close to the neural avatar, while the coarse world space distance $d_{coarse}^{world}$ could also serve as a good approximation for the world-space distance $d^{world}$ because the parametric human model is aligned to the neural avatar's geometry.

\input{tables/comparison_full}
\input{figures/more_qualitative_results}

%% file: figures/comparison_mono.tex
\begin{figure*}[ht]
    \centering
    \includegraphics[width=1.0\textwidth]{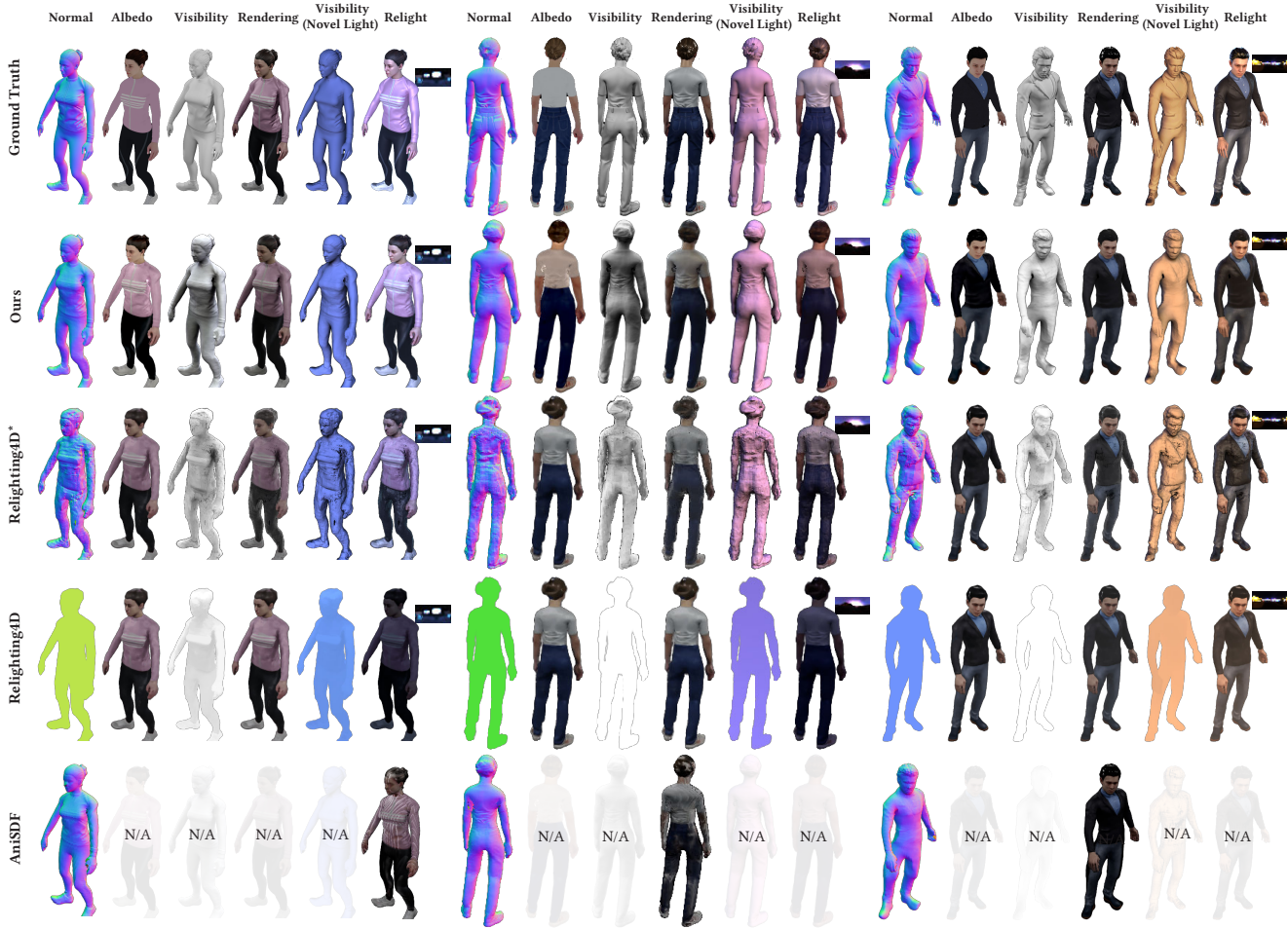}
    \caption{\textbf{Qualitative Comparison of our method and baselines.} 
    We present qualitative comparisons between our method and various baselines given only monocular input. The brightness difference between the ground truth and our method comes of the inherant scale ambiguity of recovering the environment lighting and albedo at the same time. The normal and visibility MLPs of Relighting4D \cite{chen2022relighting4d} did not successfully capture the complex motion of the human avatar when only given monocular video as input thus producing overly smooth normal map that looks like a single-colored image. Note that NeRFactor is only trained on 1 frame and all 10 input views.} 
    \label{fig:comparison_mono}
\end{figure*}

%% file: figures/ablation_hdq_distance.tex
\begin{figure}[ht]
    \centering
    \includegraphics[width=0.45\textwidth]{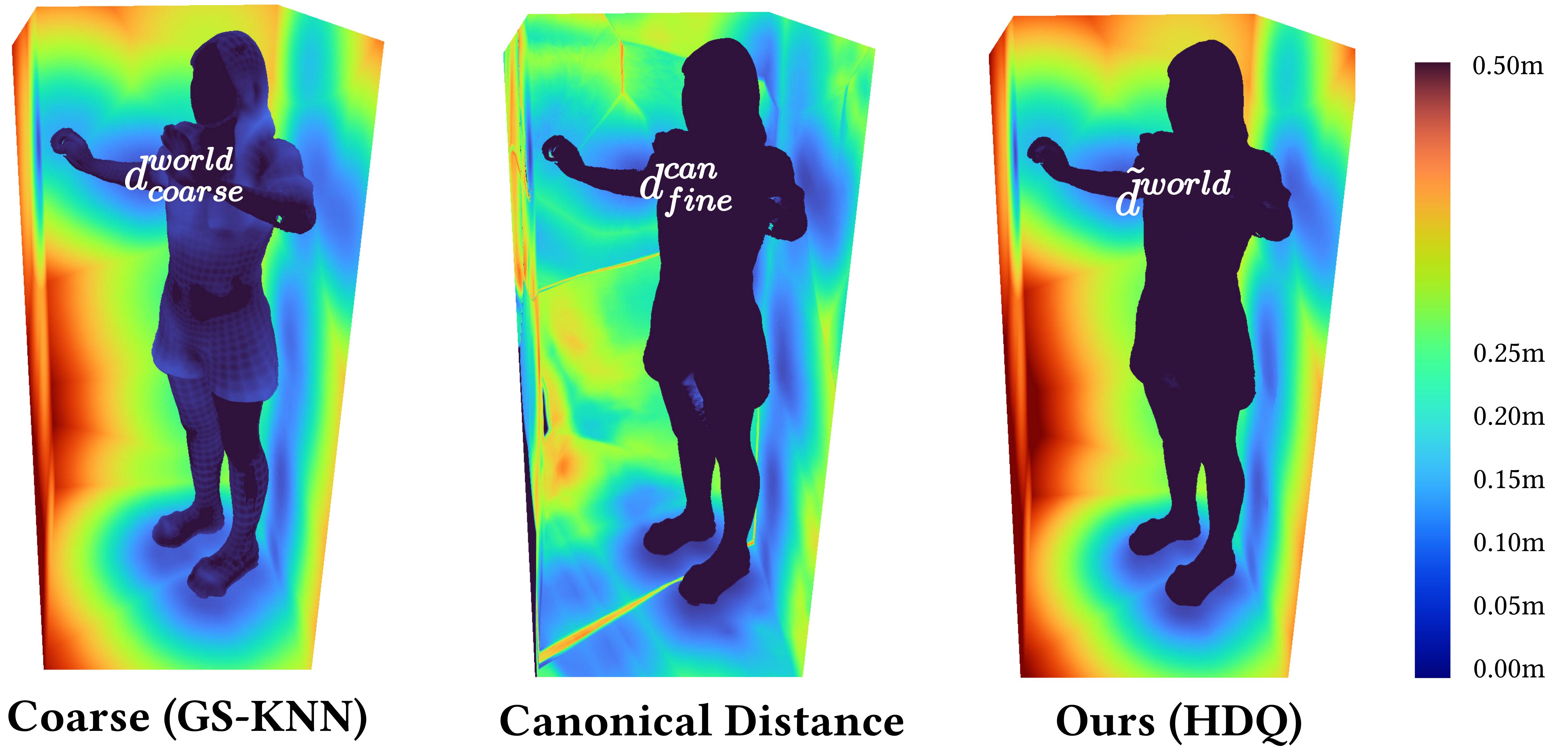}
    \caption{
        \textbf{Visualization of hierarchically queried distance}. Hot colors indicate large distance values. We visualize $d^{world}_{coarse}$, $d^{can}_{fine}$ and $\tilde{d}^{world}$ on (a) surface intersection points, which should be zero everywhere, and (b) bounding box of the human model, which should reflect the real world space SDF. Note that the geometry is fixed for the comparison since we want to visualize the distance values on the correct geometry. The coarse distance $d^{world}_{coarse}$ of near surfaces points is not strictly zero (indicated by the light blue regions on the body) due to misalignment between SMPL-H \cite{romero2017embodied} and the neural avatar, and $d^{can}_{fine}$ is incorrect in world space since it is defined by the canonical space network $S$. This leads to both of them reporting errornous geometry for the relighting task as seen in Figure~\ref{fig:hdq_geo}. \textbf{In contrast to both, our proposed HDQ algorithm is well behaved both when near and far from the geometry.}
    }
    \label{fig:ablation_hdq_distance}
\end{figure}

%% file: tables/ablation_dfss.tex
\begin{table}[ht]
    \caption{
        \textbf{Ablation on other hyperparameters used in the model.} $N_{st}$ and $N_{st}^{vis}$ denote the number of sphere tracing steps for pixel-surface intersection and soft light visibility estimation respectively. $K$ is the number of vertices sampled in the inverse warping process of GS-KNN and $N_{s}$ is the number canonical material samples for volume rendering albedo $\alpha$ and roughness $\gamma$.
    }
    \label{tab:ablation_dfss}
    \centering
    \addtolength{\tabcolsep}{-1pt}
    \resizebox{1.0\linewidth}{!}{

        \begin{tabular}{lcccccc}
            \toprule
            \multicolumn{1}{c}{} & \multicolumn{3}{c}{\textbf{Relighting}} & \multicolumn{3}{c}{\textbf{Visibility}}                                                                           \\
                                 & PSNR$\uparrow$                          & SSIM$\uparrow$                          & LPIPS$\downarrow$ & PSNR$\uparrow$ & SSIM$\uparrow$ & LPIPS$\downarrow$ \\
            \midrule
            $N_{st} = 32$        & 20.84                                   & 0.810                                   & 0.201             & \textbf{18.45} & \textbf{0.798} & 0.193             \\
            $N_{st} = 16$        & \textbf{21.08}                          & \textbf{0.815}                          & \textbf{0.197}    & 18.19          & \textbf{0.798} & \textbf{0.187}    \\
            $N_{st} = 8$         & 20.67                                   & 0.773                                   & 0.242             & 17.50          & 0.726          & 0.266             \\
            $N_{st} = 4$         & 20.43                                   & 0.680                                   & 0.350             & 15.88          & 0.578          & 0.402             \\
            $N_{st} = 2$         & 19.22                                   & 0.589                                   & 0.428             & 13.38          & 0.494          & 0.465             \\
            \midrule
            $N^{vis}_{st} = 8$   & 20.99                                   & 0.811                                   & \textbf{0.196}    & 18.27          & 0.802          & 0.187             \\
            $N^{vis}_{st} = 4$   & \textbf{21.08}                          & \textbf{0.815}                          & 0.197             & 18.19          & 0.798          & 0.187             \\
            $N^{vis}_{st} = 2$   & 20.71                                   & 0.812                                   & 0.199             & \textbf{18.63} & \textbf{0.806} & \textbf{0.184}    \\
            $N^{vis}_{st} = 1$   & 20.55                                   & 0.804                                   & 0.202             & 18.49          & 0.804          & 0.186             \\
            \midrule
            $K = 10$             & \textbf{21.08}                          & \textbf{0.815}                          & \textbf{0.197}    & 18.19          & 0.798          & \textbf{0.187}    \\
            $K = 5$              & 20.76                                   & 0.810                                   & 0.203             & 18.17          & 0.795          & 0.195             \\
            $K = 2$              & 20.78                                   & 0.812                                   & 0.198             & \textbf{18.42} & \textbf{0.803} & 0.188             \\
            \midrule
            $N_s = 10$           & 20.65                                   & 0.813                                   & 0.199             & 17.98          & 0.800          & 0.195             \\
            $N_s = 5$            & 20.80                                   & 0.811                                   & \textbf{0.202}    & 18.42          & 0.797          & 0.194             \\
            $N_s = 3$            & \textbf{21.08}                          & \textbf{0.815}                          & 0.197             & 18.19          & 0.798          & 0.187             \\
            $N_s = 1$            & 20.81                                   & 0.813                                   & 0.200             & \textbf{18.48} & \textbf{0.804} & \textbf{0.183}    \\
            \bottomrule
        \end{tabular}
    }
    \addtolength{\tabcolsep}{1pt}

\end{table}

%% file: figures/ablation_dfss.tex
\begin{figure}[ht]
    \centering
    \includegraphics[width=0.45\textwidth]{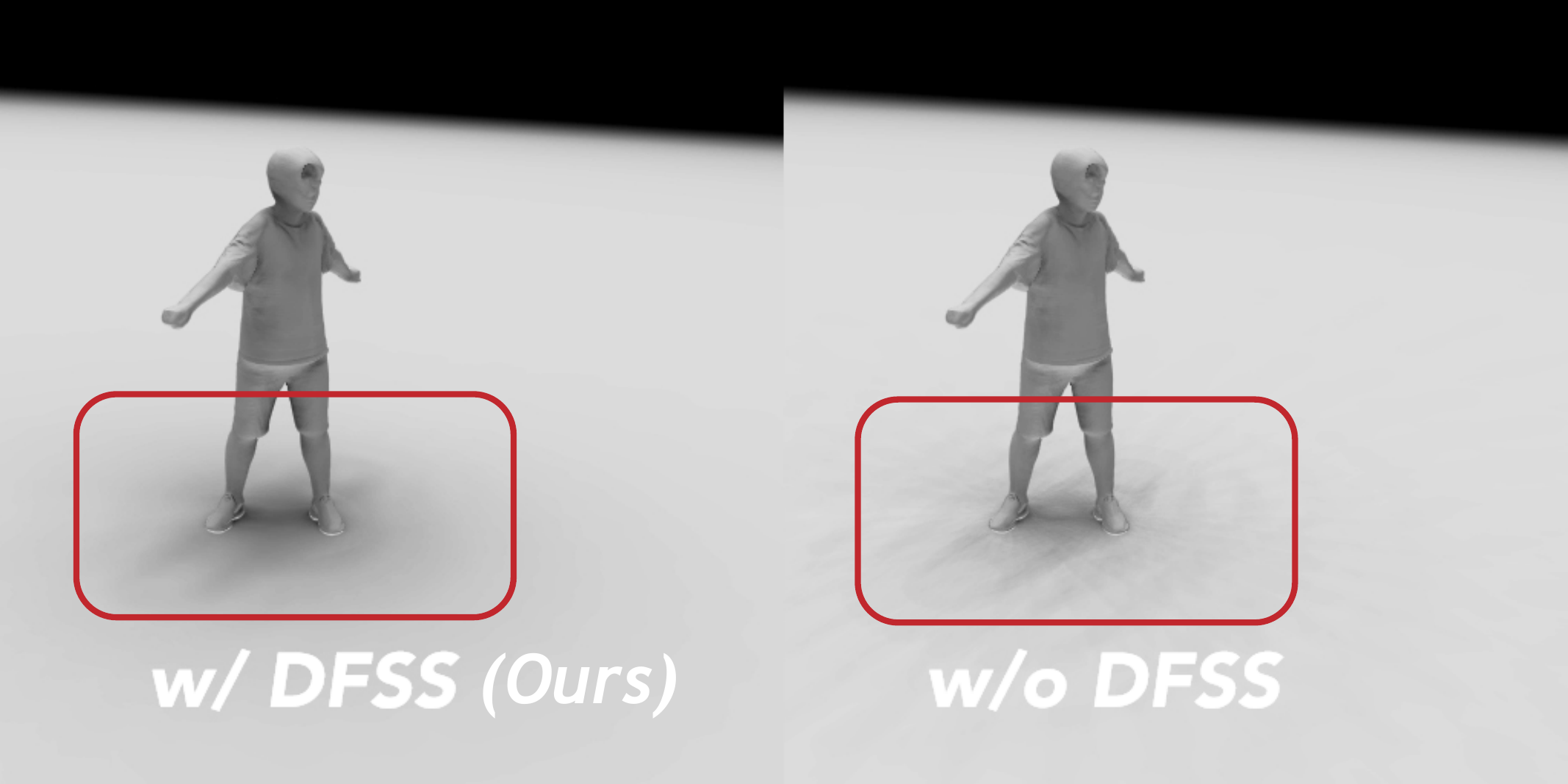}
    \caption{
        \textbf{Comparison of light visibility estimation algorithm}. We visualize the ambient occlusion result of different shadow algorithms by casting shadows from a dome lighting where the upper half of the environment map is set to a constant brightnerss. We use $N_{st}=16$ sphere tracing iterations to generate shadows on the ground for ``w/o DFSS'' and ``w/ DFSS (ours)''. The ``w/o DFSS'' variant takes the binary light occlusion mask as the visibility term, resulting in unnatural hard shadow. \textbf{``w/ DFSS (ours)'' generates realistic soft shadows by incorporating \cite{parker1998single} onto neural SDF produced by ours Hierarchical Distance Query.}
    }
    \label{fig:ablation_dfss}
\end{figure}

%% file: tables/ablation_sdf.tex
\begin{table}[ht]

    \caption{
        \textbf{Ablation study on the accuracy of the proposed HDQ algorithm and sphere tracing procedure.}
        We provide comparison on the average absolute distance (denoted $\mathrm{abs}(\mathrm{SDF}(\boldsymbol{x}_s))$) of the computed pixel-surface intersection on four variants of the HDQ algorithm to compare its accuracy. Our method achieves an average accuracy of $0.00017$m while the other variants exhibit at least two orders of magnitude higher error.
    }

    \label{tab:ablation_sdf}
    \centering
    \addtolength{\tabcolsep}{-2pt}
    \resizebox{1.0\linewidth}{!}{

    \begin{tabular}{ccccccc}
        \toprule
                                                       & \textbf{ours} & w/o HDQ & w/o $d_{coarse}^{world}$ & w/o $d_{fine}^{can}$ \\
        \midrule
        $\mathrm{abs}(\mathrm{SDF}(\boldsymbol{x}_s))$ & 0.00017       & 0.20462 & 0.06774                  & 0.00961              \\
        \bottomrule
    \end{tabular}
    }
    \addtolength{\tabcolsep}{1pt}

\end{table}

%% file: figures/hdq_geo.tex
\begin{figure}
    \centering
    \includegraphics[width=0.45\textwidth]{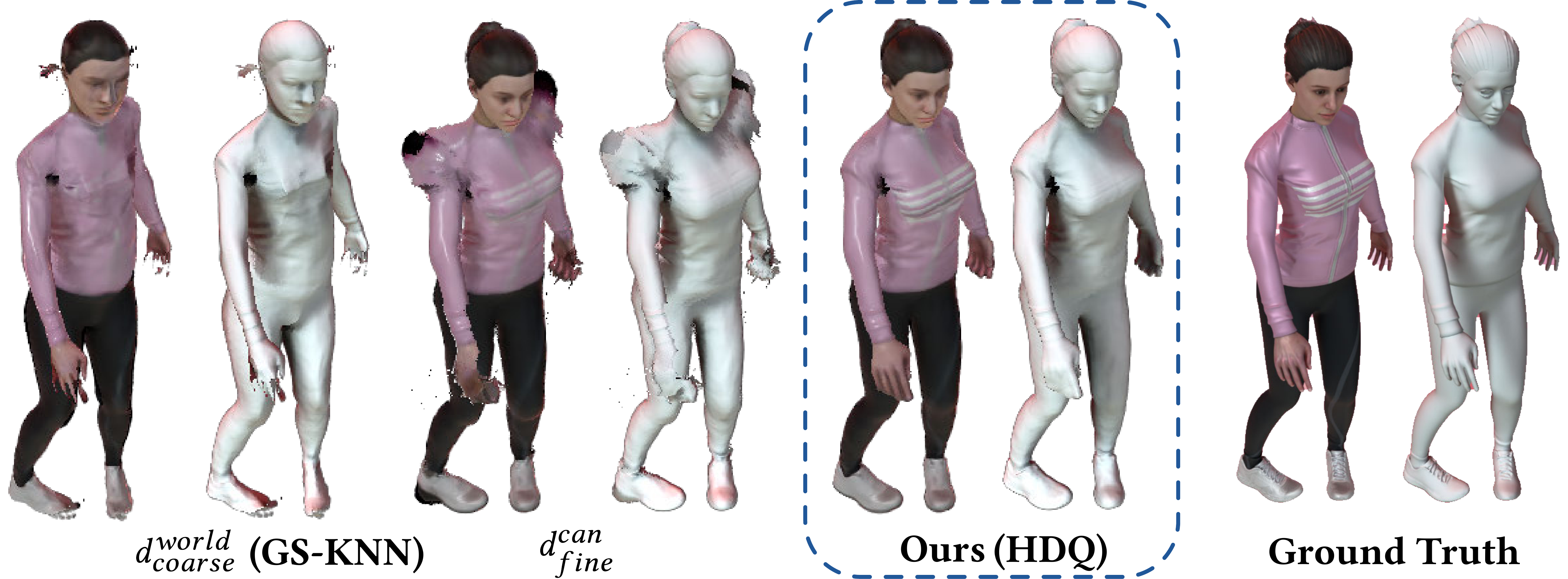}
    \caption{
        \textbf{Geometry quality of GS-KNN and HDQ}. Performing sphere tracing using only the canonical distance $d_{fine}^{can}$ or coarse world distance $d_{coarse}^{world}$ (GS-KNN) results in incorrect surface intersection $\boldsymbol{x}_s$ and soft visibility $p_s$, while tracing with our proposed Hierarchical Distance Query produces correct results.
    }
    \label{fig:hdq_geo}
\end{figure}

%% file: tables/ablation_cutoff.tex
\begin{table}[ht]
    \caption{
    \textbf{Sensitivity study and runtime analysis on the cut-off value.} The frame time and the rendering quality is roughly linear to the cut-off value up to a certain point, after which only diminishing returns can be observed by increasing the cut-off. However, a too small value may result in incorrect surface intersection and visibility estimation, leading to degraded quality. Thus we choose the minimum cut-off value without a visible quality degredation ($\tilde{T}_d = 0.1, \tilde{T}_d^{vis} = 0.025$) as the default one.
    }
    \label{tab:ablatin_cutoff}
    \centering
    \addtolength{\tabcolsep}{-5pt}

    \resizebox{1.0\linewidth}{!}{
        \begin{tabular}{lccccccc}
            \toprule
            \multicolumn{1}{c}{}                             & \multicolumn{1}{c}{}   & \multicolumn{3}{c}{\textbf{Relighting}} & \multicolumn{3}{c}{\textbf{Visibility}}                                                                           \\
                                                             & Frame Time$\downarrow$ & PSNR$\uparrow$                          & SSIM$\uparrow$                          & LPIPS$\downarrow$ & PSNR$\uparrow$ & SSIM$\uparrow$ & LPIPS$\downarrow$ \\
            \midrule
            $\tilde{T}_d = 2.0, \tilde{T}_d^{vis} = 0.5$     & 7.659                  & \textbf{21.11}                          & \textbf{0.815}                          & 0.200             & 18.10          & 0.795          & 0.193             \\
            $\tilde{T}_d = 1.0, \tilde{T}_d^{vis} = 0.25$    & 7.675                  & \textbf{21.11}                          & \textbf{0.815}                          & 0.200             & 18.17          & 0.796          & 0.191             \\
            $\tilde{T}_d = 0.5, \tilde{T}_d^{vis} = 0.125$   & 7.320                  & 21.10                                   & \textbf{0.815}                          & 0.199             & 18.17          & 0.797          & 0.189             \\
            $\tilde{T}_d = 0.1, \tilde{T}_d^{vis} = 0.025$   & 4.524                  & 21.08                                   & \textbf{0.815}                          & \textbf{0.197}    & \textbf{18.19} & \textbf{0.798} & \textbf{0.187}    \\
            $\tilde{T}_d = 0.05, \tilde{T}_d^{vis} = 0.0125$ & 2.631                  & 21.04                                   & 0.814                                   & 0.202             & 17.83          & 0.790          & 0.197             \\
            $\tilde{T}_d = 0.01, \tilde{T}_d^{vis} = 0.0025$ & \textbf{0.976}         & 19.50                                   & 0.733                                   & 0.300             & 15.82          & 0.697          & 0.304             \\
            \bottomrule
        \end{tabular}
    }
    \addtolength{\tabcolsep}{5pt}

\end{table}

%% file: tables/dataset_setting.tex
\begin{table}[ht]
    \caption{
        \textbf{Specific training settings.} $N_{view}$ and $N_{frame}$ represent the number of training views and training frames respectively. $N_{view}^{total}$ denotes the total number of views of the dataset and $N_{frame}^{total}$ denotes the total number of frames.
    }
    \label{tab:training_settings}
    \centering

    \addtolength{\tabcolsep}{-2pt}
    \resizebox{0.9\linewidth}{!}{

        \begin{tabular}{p{0.225\linewidth} ccc} %
            \toprule
            Dataset                               & Character        & $N_{view}$/$N_{view}^{total}$ & $N_{frame}$/$N_{frame}^{total}$ \\
            \midrule
            \multirow{4}{*}{\textit{MobileStage}} & \textit{dark}    & \textbf{12} / 36              & 1600 / 2000                     \\
                                                  & \textit{purple}  & \textbf{12} / 36              & 600  / 700                      \\
                                                  & \textit{black}   & \textbf{12} / 36              & 300  / 400                      \\
                                                  & \textit{white}   & \textbf{12} / 36              & 300  / 600                      \\
            \midrule
                                                  & \textit{jody}    & \textbf{1} / 20               & 100  / 100                      \\
            \textit{SynthHuman}                   & \textit{josh}    & \textbf{1} / 20               & 100  / 100                      \\
            (\textbf{monocular})                  & \textit{megan}   & \textbf{1} / 20               & 100  / 100                      \\
                                                  & \textit{leonard} & \textbf{1} / 20               & 100  / 100                      \\
            \bottomrule
        \end{tabular}
    }
    \addtolength{\tabcolsep}{2pt}

\end{table}

%% file: figures/ablation_vis_albedo.tex
\begin{figure}[ht]
    \centering
    \includegraphics[width=0.25\textwidth]{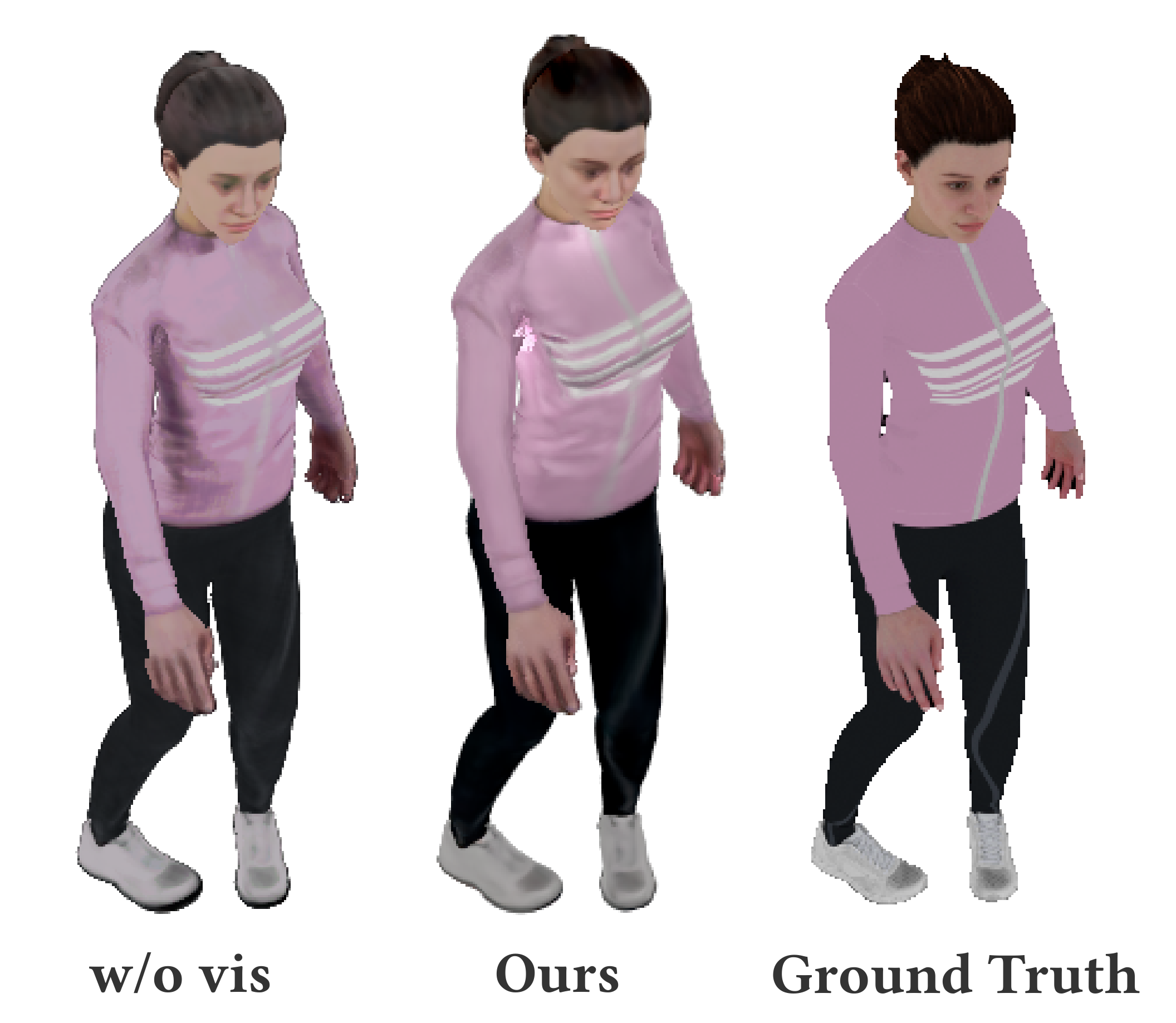}
    \caption{
        \textbf{Modeling visibility helps reconstructing diffuse albedo.} The ``w/o vis'' variant trained without computing light visibility bakes shadows onto the diffuse albedo, while our approach reconstructs high quality albedo.
    }
    \label{fig:ablation_vis_albedo}
\end{figure}

%% file: algos/geo_st.tex
\begin{algorithm}
    \small
    \LinesNumbered
    \DontPrintSemicolon
    \caption{Surface intersection sphere tracing.}
    \label{alg:surface_intersection_sphere_tracing}

    \KwIn{$\boldsymbol{o}$, $\boldsymbol{d}$, $N_{st}$, $n$, $f$}
    \KwOut{$\boldsymbol{x}_{s}$}

    $t \gets n$ \tcp*[rh]{Start from near plane}
    $t_s \gets f$ \tcp*[rh]{Surface intersection depth}
    $d_{0} \gets \inf$ \tcp*[rh]{Previous closest distance}
    $d_{1} \gets \inf$ \tcp*[rh]{Current closest distance}
    $d_{c} \gets \inf$ \tcp*[rh]{All time closest distance}
    $d_{t} \gets \inf$ \tcp*[rh]{Update of $t$ during iteration}
    \For{$i \gets 0$ \KwTo $N_{st}$}{

        \tcp*[l]{Hierarchical distance query}
        $d_{0}, d_{1} \gets d_{1}, \tilde{d}^{world}(\boldsymbol{o} + t \boldsymbol{d})$ \;
        $d^{abs}_{0}, d^{abs}_{1} \gets \mathrm{abs}(d_{0}), \mathrm{abs}(d_{1})$ \;

        \tcp*[l]{Update closest distance and intersection}
        \If{$d^{abs}_{1} < d_{c}$} {
            $d_{c} \gets d^{abs}_{1}$ \;
            $t_s \gets t$ \;
        }

        \tcp*[l]{Linear interpolation upon sign change of SDF}
        \If{$\mathrm{sign}(d_{0}) \neq \mathrm{sign}(d_{1})$} {
            $t_s \gets t - d_t \frac{ d^{abs}_{1}}{d^{abs}_{0} + d^{abs}_{1}}$  \;
        }
        \tcp*[l]{Prepare for next iteration}
        $d_t \gets d + o$ \;
        $t \gets t + d_t$ \;
        \tcp*[l]{Constrain $t$ to be within near and far plane}
        $t \gets \mathrm{min}(t, f)$ \;
        $t \gets \mathrm{max}(t, n)$ \;
    }
    \Return $\boldsymbol{o} + t_s \boldsymbol{d}$ \;
\end{algorithm}

%% file: algos/dfss_st.tex
\begin{algorithm}
    \small
    \LinesNumbered
    \DontPrintSemicolon
    \caption{Soft light visibility sphere tracing.}
    \label{alg:soft_light_visibility_sphere_tracing}

    \KwIn{$\boldsymbol{o}$, $\boldsymbol{d}$, $N^{vis}_{st}$, $n$, $f$, $a$}
    \KwOut{$p$}

    $t \gets n$ \tcp*[rh]{Start from near plane}
    $d_{0} \gets \inf$ \tcp*[rh]{Previous closest distance}
    $d_{1} \gets \inf$ \tcp*[rh]{Current closest distance}
    $d_{t} \gets \inf$ \tcp*[rh]{Update of $t$ during iteration}
    $p_{s} \gets 1$ \tcp*[rh]{Soft shadow penumbra coefficient}
    $R_{s} \gets \sqrt{\frac{a}{\pi}}$ \tcp*[rh]{Per solid angle $a = \pi R_{s}^2$}

    \For{$i \gets 0$ \KwTo $N^{vis}_{st}$}{

        \tcp*[l]{Hierarchical distance query}
        $d_{0}, d_{1} \gets d_{1}, \tilde{d}^{world}(\boldsymbol{o} + t \boldsymbol{d})$ \;

        \tcp*[l]{Compute penumbra coefficient}
        $p_{s} \gets \mathrm{min}(p_{s}, \frac{\mathrm{max}(d_1, 0)}{2 t R_{s}})$ \;

        \tcp*[l]{Prepare for next iteration}
        $d_t \gets d + o$ \;
        $t \gets t + d_t$ \;
        \tcp*[l]{Constrain $t$ to be within near and far plane}
        $t \gets \mathrm{min}(t, f)$ \;
        $t \gets \mathrm{max}(t, n)$ \;
    }
    \Return $p_{s}$ \;
\end{algorithm}

%% file: figures/network_structure.tex
\begin{figure*}[ht]
    \centering
    \includegraphics[width=0.85\textwidth]{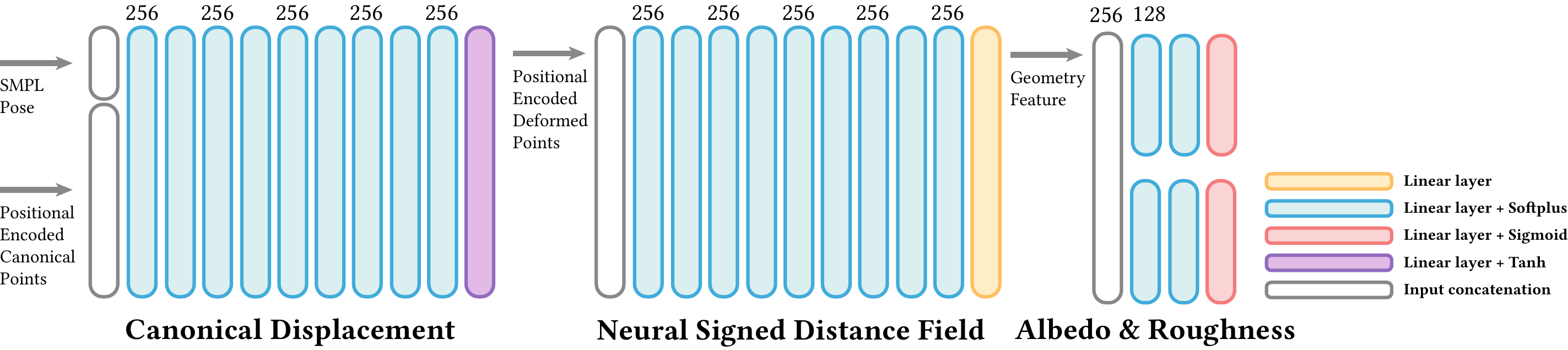}
    \caption{\textbf{Detailed network structure.}}
    \label{fig:network_structure}
\end{figure*}

%% file: tables/comparison_full.tex
\begin{table*}[ht]
    \caption{
        \textbf{Addtional quantitative comparison on all characters of SyntheticHuman++.} We report the result of NeRFactor and NeRFactor* on the first frame of each sequence. NeRFactor \cite{zhang2021nerfactor} and NeRFactor* failed to converge on the first frame of \textit{megan} and \textit{leonard}.
    }
    \label{tab:comparison_full}
    \centering

    \addtolength{\tabcolsep}{-2pt}
    \resizebox{0.9\linewidth}{!}{

        \begin{tabular}{llcccccccccc} %
            \toprule
            \multicolumn{2}{c}{}              & \multicolumn{1}{c}{\textbf{Normal}} & \multicolumn{3}{c}{\textbf{Diffuse Albedo}} & \multicolumn{3}{c}{\textbf{Relighting}} & \multicolumn{3}{c}{\textbf{Visibility}}                                                                                                                                 \\
            Character                         & Method                              & Degree $\downarrow$                         & PSNR$\uparrow$                          & SSIM$\uparrow$                          & LPIPS$\downarrow$ & PSNR$\uparrow$ & SSIM$\uparrow$ & LPIPS$\downarrow$ & PSNR$\uparrow$ & SSIM$\uparrow$ & LPIPS$\downarrow$ \\
            \midrule
            \multirow{4}{*}{\textit{jody}}    & Ours                                & 10.98                                       & 25.88                                   & 0.925                                   & 0.144             & 21.57          & 0.853          & 0.168             & 20.53          & 0.869          & 0.142             \\
                                              & Relighting4D*                       & 27.65                                       & 16.40                                   & 0.827                                   & 0.229             & 20.77          & 0.826          & 0.205             & 15.47          & 0.785          & 0.241             \\
                                              & Relighting4D                        & 76.73                                       & 16.40                                   & 0.827                                   & 0.229             & 17.25          & 0.752          & 0.249             & 9.51           & 0.604          & 0.313             \\
                                              & NeRFactor*                          & 33.19                                       & 17.74                                   & 0.860                                   & 0.201             & 21.66          & 0.867          & 0.224             & 16.12          & 0.752          & 0.283             \\
                                              & NeRFactor                           & 79.26                                       & 17.74                                   & 0.860                                   & 0.201             & 20.31          & 0.862          & 0.181             & 11.78          & 0.755          & 0.227             \\
                                              & AniSDF                              & 13.61                                       & 16.02                                   & 0.815                                   & 0.230             & 19.63          & 0.818          & 0.223             & -              & -              & -                 \\
            \multirow{4}{*}{\textit{josh}}    & Ours                                & 12.88                                       & 34.46                                   & 0.968                                   & 0.072             & 24.72          & 0.897          & 0.206             & 20.04          & 0.858          & 0.149             \\
                                              & Relighting4D*                       & 26.56                                       & 29.32                                   & 0.917                                   & 0.153             & 24.48          & 0.881          & 0.229             & 16.02          & 0.790          & 0.233             \\
                                              & Relighting4D                        & 101.0                                       & 29.31                                   & 0.917                                   & 0.153             & 23.86          & 0.830          & 0.273             & 3.25           & 0.554          & 0.363             \\
                                              & NeRFactor*                          & 35.39                                       & 29.14                                   & 0.926                                   & 0.167             & 24.68          & 0.895          & 0.228             & 14.85          & 0.729          & 0.291             \\
                                              & NeRFactor                           & 24.57                                       & 29.14                                   & 0.926                                   & 0.167             & 24.65          & 0.896          & 0.217             & 15.96          & 0.791          & 0.245             \\
                                              & AniSDF                              & 16.34                                       & 30.28                                   & 0.924                                   & 0.164             & 20.76          & 0.843          & 0.260             & -              & -              & -                 \\
            \multirow{4}{*}{\textit{megan}}   & Ours                                & 13.09                                       & 22.67                                   & 0.873                                   & 0.198             & 18.55          & 0.790          & 0.234             & 19.69          & 0.814          & 0.167             \\
                                              & Relighting4D*                       & 31.98                                       & 22.91                                   & 0.857                                   & 0.223             & 17.25          & 0.742          & 0.273             & 14.83          & 0.713          & 0.278             \\
                                              & Relighting4D                        & 98.05                                       & 22.93                                   & 0.857                                   & 0.222             & 17.26          & 0.671          & 0.307             & 5.47           & 0.364          & 0.433             \\
                                              & NeRFactor*                          & -                                           & 15.11                                   & 0.569                                   & 0.394             & 14.06          & 0.490          & 0.472             & 7.41           & 0.341          & 0.538             \\
                                              & NeRFactor                           & -                                           & 15.11                                   & 0.569                                   & 0.394             & 14.06          & 0.490          & 0.472             & 7.41           & 0.341          & 0.538             \\
                                              & AniSDF                              & 15.18                                       & 16.91                                   & 0.788                                   & 0.258             & 11.51          & 0.686          & 0.316             & -              & -              & -                 \\
            \multirow{4}{*}{\textit{leonard}} & Ours                                & 12.81                                       & 33.04                                   & 0.968                                   & 0.064             & 25.92          & 0.903          & 0.218             & 20.53          & 0.852          & 0.160             \\
                                              & Relighting4D*                       & 31.33                                       & 30.18                                   & 0.938                                   & 0.129             & 26.03          & 0.891          & 0.242             & 14.57          & 0.766          & 0.255             \\
                                              & Relighting4D                        & 99.54                                       & 30.18                                   & 0.938                                   & 0.130             & 25.14          & 0.845          & 0.273             & 3.24           & 0.533          & 0.388             \\
                                              & NeRFactor*                          & -                                           & 26.93                                   & 0.912                                   & 0.143             & 23.77          & 0.779          & 0.326             & 7.09           & 0.500          & 0.435             \\
                                              & NeRFactor                           & -                                           & 26.93                                   & 0.912                                   & 0.143             & 23.77          & 0.779          & 0.326             & 7.09           & 0.500          & 0.435             \\
                                              & AniSDF                              & 13.75                                       & 25.31                                   & 0.920                                   & 0.155             & 18.30          & 0.849          & 0.246             & -              & -              & -                 \\

            \bottomrule
        \end{tabular}
    }
    \addtolength{\tabcolsep}{2pt}

\end{table*}

\begin{table*}[ht]
    \caption{
        \textbf{Addtional quantitative comparison on all characters of SyntheticHuman++ under the monocular setting.}
    }
    \label{tab:comparison_mono}
    \centering

    \addtolength{\tabcolsep}{-2pt}
    \resizebox{0.9\linewidth}{!}{

        \begin{tabular}{cccccccccccc} %
            \toprule
            \multicolumn{2}{c}{}              & \multicolumn{1}{c}{\textbf{Normal}} & \multicolumn{3}{c}{\textbf{Diffuse Albedo}} & \multicolumn{3}{c}{\textbf{Relighting}} & \multicolumn{3}{c}{\textbf{Visibility}}                                                                                                                                 \\
            Character                         & Method                              & Degree $\downarrow$                         & PSNR$\uparrow$                          & SSIM$\uparrow$                          & LPIPS$\downarrow$ & PSNR$\uparrow$ & SSIM$\uparrow$ & LPIPS$\downarrow$ & PSNR$\uparrow$ & SSIM$\uparrow$ & LPIPS$\downarrow$ \\
            \midrule
            \multirow{4}{*}{\textit{jody}}    & Ours                                & 16.83                                       & 24.09                                   & 0.877                                   & 0.180             & 21.08          & 0.815          & 0.197             & 18.19          & 0.798          & 0.187             \\
                                              & Relighting4D*                       & 25.16                                       & 19.57                                   & 0.799                                   & 0.261             & 20.79          & 0.766          & 0.245             & 17.31          & 0.736          & 0.273             \\
                                              & Relighting4D                        & 80.80                                       & 19.57                                   & 0.799                                   & 0.262             & 20.28          & 0.778          & 0.233             & 16.16          & 0.753          & 0.278             \\
                                              & AniSDF                              & 18.65                                       & 16.25                                   & 0.752                                   & 0.284             & 18.65          & 0.746          & 0.274             & -              & -              & -                 \\
            \multirow{4}{*}{\textit{josh}}    & Ours                                & 20.09                                       & 26.67                                   & 0.905                                   & 0.142             & 24.52          & 0.873          & 0.224             & 17.46          & 0.761          & 0.222             \\
                                              & Relighting4D*                       & 24.93                                       & 29.61                                   & 0.919                                   & 0.165             & 23.95          & 0.863          & 0.240             & 17.29          & 0.737          & 0.268             \\
                                              & Relighting4D                        & 81.23                                       & 29.61                                   & 0.919                                   & 0.165             & 24.06          & 0.861          & 0.269             & 15.88          & 0.750          & 0.297             \\
                                              & AniSDF                              & 21.91                                       & 26.54                                   & 0.879                                   & 0.226             & 19.88          & 0.790          & 0.294             & -              & -              & -                 \\
            \multirow{4}{*}{\textit{megan}}   & Ours                                & 19.64                                       & 18.63                                   & 0.796                                   & 0.262             & 18.33          & 0.742          & 0.246             & 17.70          & 0.707          & 0.222             \\
                                              & Relighting4D*                       & 29.51                                       & 22.38                                   & 0.798                                   & 0.286             & 17.21          & 0.696          & 0.285             & 16.29          & 0.640          & 0.316             \\
                                              & Relighting4D                        & 83.40                                       & 22.38                                   & 0.798                                   & 0.286             & 17.59          & 0.705          & 0.299             & 16.30          & 0.648          & 0.354             \\
                                              & AniSDF                              & 19.80                                       & 26.29                                   & 0.905                                   & 0.179             & 20.77          & 0.825          & 0.274             & -              & -              & -                 \\
            \multirow{4}{*}{\textit{leonard}} & Ours                                & 18.30                                       & 24.28                                   & 0.912                                   & 0.121             & 25.88          & 0.893          & 0.231             & 18.45          & 0.779          & 0.215             \\
                                              & Relighting4D*                       & 25.08                                       & 29.90                                   & 0.940                                   & 0.130             & 25.28          & 0.884          & 0.245             & 17.50          & 0.726          & 0.286             \\
                                              & Relighting4D                        & 81.55                                       & 29.90                                   & 0.940                                   & 0.130             & 25.48          & 0.879          & 0.270             & 16.40          & 0.753          & 0.280             \\
                                              & AniSDF                              & 21.08                                       & 16.98                                   & 0.710                                   & 0.332             & 13.85          & 0.619          & 0.346             & -              & -              & -                 \\

            \bottomrule
        \end{tabular}
    }
    \addtolength{\tabcolsep}{2pt}

\end{table*}

%% file: figures/more_qualitative_results.tex
\begin{figure*}[ht]
    \centering
    \includegraphics[width=0.95\textwidth]{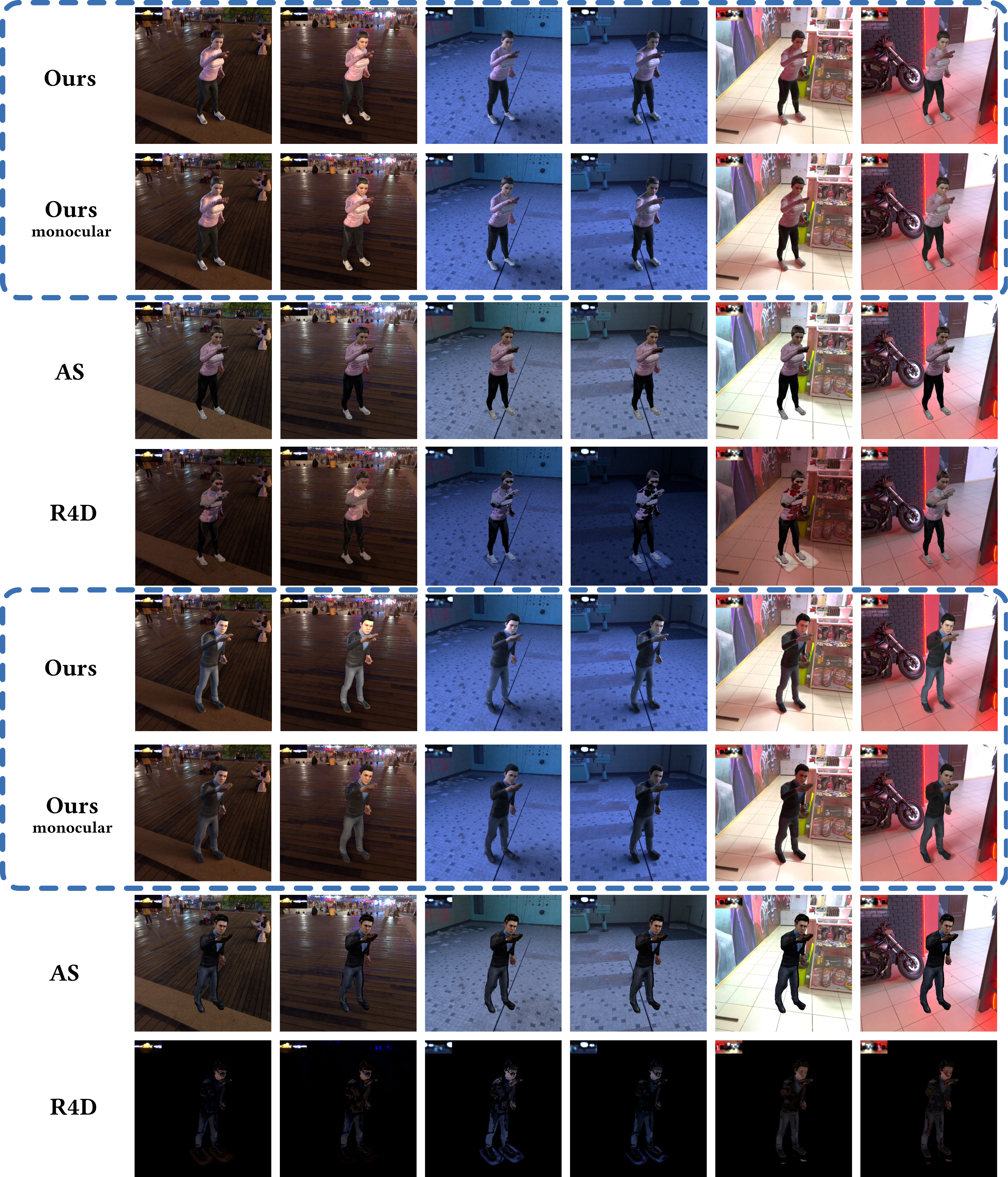}
    \caption{\textbf{More novel pose relighting results on the SyntheticHuman dataset.} ``AS'' denotes AniSDF \cite{peng2022animatable}. ``R4D'' denotes Relighting4D \cite{chen2022relighting4d}. All compared relighting methods compute shadows on the ground plane. The visibility MLP of \cite{chen2022relighting4d} failed to generalize to far-from-human points.}
    \label{fig:synthetic_human_novel}
\end{figure*}

\begin{figure*}[ht]
    \centering
    \includegraphics[width=0.95\textwidth]{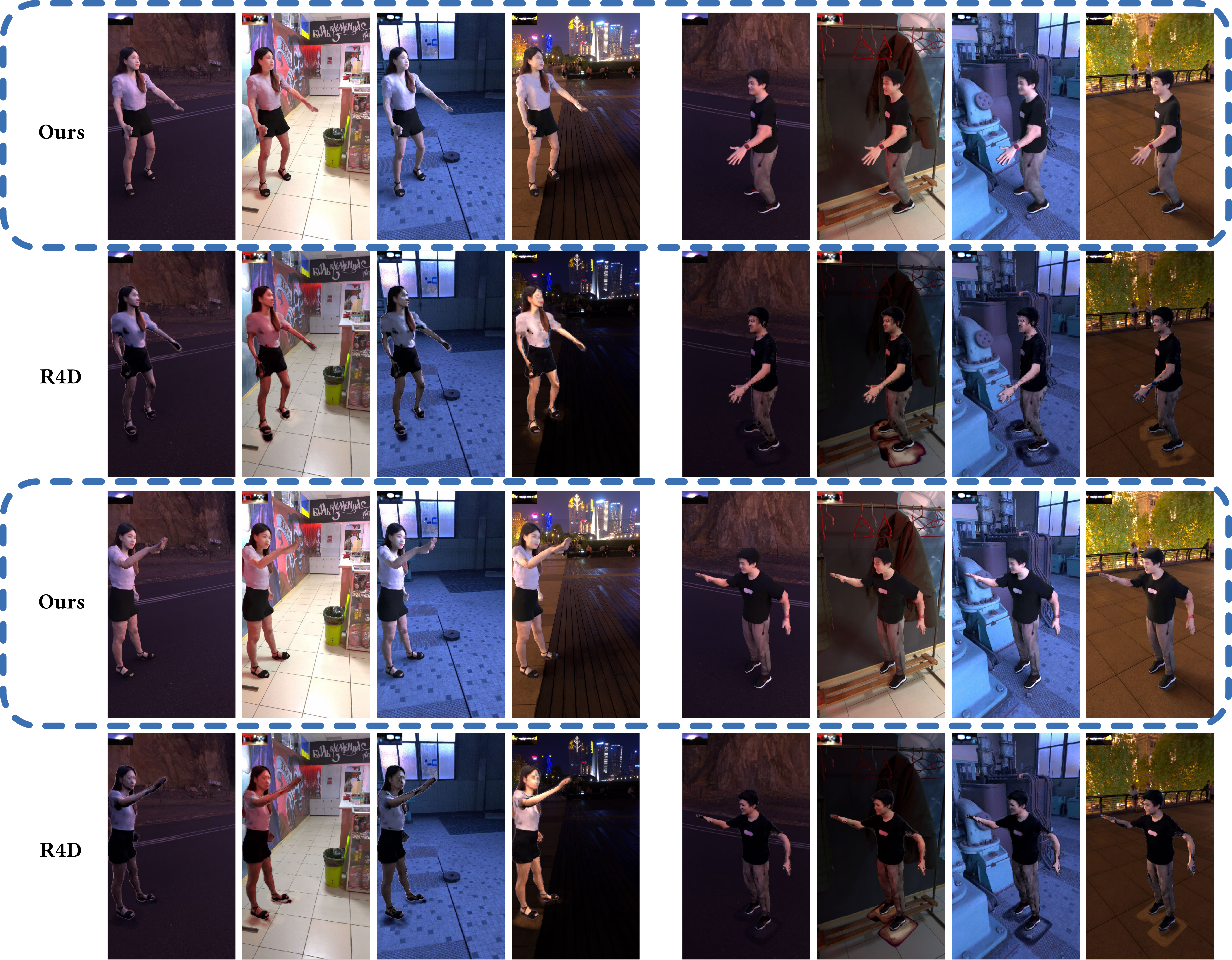}
    \caption{\textbf{More novel pose relighting results on the MobileStage dataset.} ``AS'' denotes AniSDF \cite{peng2022animatable}. ``R4D'' denotes Relighting4D \cite{chen2022relighting4d}. All compared relighting methods compute shadows on the ground plane. The visibility MLP of \cite{chen2022relighting4d} failed to generalize to far-from-human points.} 
    \label{fig:mobile_stage_novel}
\end{figure*}

%% file: main.bbl
\begin{thebibliography}{10}\itemsep=-1pt

\bibitem{easymocap}
Easymocap - make human motion capture easier.
\newblock Github, 2021.

\bibitem{polyhaven}
Poly {Haven}, 2023.

\bibitem{aaltonen2018gpu}
Sebastian Aaltonen.
\newblock Gpu-based clay simulation and ray-tracing tech in claybook.
\newblock {\em San Francisco, CA}, 2(5), 2018.

\bibitem{aitpayev2012creation}
Kairat Aitpayev and Jaafar Gaber.
\newblock Creation of 3d human avatar using kinect.
\newblock {\em Asian Transactions on Fundamentals of Electronics, Communication \& Multimedia}, 1(5):12--24, 2012.

\bibitem{alldieck2018video}
Thiemo Alldieck, Marcus Magnor, Weipeng Xu, Christian Theobalt, and Gerard Pons-Moll.
\newblock Video based reconstruction of 3d people models.
\newblock In {\em Proceedings of the IEEE Conference on Computer Vision and Pattern Recognition}, pages 8387--8397, 2018.

\bibitem{alldieck2022photorealistic}
Thiemo Alldieck, Mihai Zanfir, and Cristian Sminchisescu.
\newblock Photorealistic monocular 3d reconstruction of humans wearing clothing.
\newblock In {\em Proceedings of the IEEE/CVF Conference on Computer Vision and Pattern Recognition}, pages 1506--1515, 2022.

\bibitem{ban2019area}
R{\'o}bert B{\'a}n, Csaba B{\'a}lint, and G{\'a}bor Valasek.
\newblock Area lights in signed distance function scenes.
\newblock In {\em Eurographics (Short Papers)}, pages 85--88, 2019.

\bibitem{bi2021deep}
Sai Bi, Stephen Lombardi, Shunsuke Saito, Tomas Simon, Shih-En Wei, Kevyn Mcphail, Ravi Ramamoorthi, Yaser Sheikh, and Jason Saragih.
\newblock Deep relightable appearance models for animatable faces.
\newblock {\em ACM Transactions on Graphics (TOG)}, 40(4):1--15, 2021.

\bibitem{bogo2015detailed}
Federica Bogo, Michael~J Black, Matthew Loper, and Javier Romero.
\newblock Detailed full-body reconstructions of moving people from monocular rgb-d sequences.
\newblock In {\em Proceedings of the IEEE international conference on computer vision}, pages 2300--2308, 2015.

\bibitem{boss2021nerd}
Mark Boss, Raphael Braun, Varun Jampani, Jonathan~T Barron, Ce Liu, and Hendrik Lensch.
\newblock Nerd: Neural reflectance decomposition from image collections.
\newblock In {\em Proceedings of the IEEE/CVF International Conference on Computer Vision}, pages 12684--12694, 2021.

\bibitem{boss2022samurai}
Mark Boss, Andreas Engelhardt, Abhishek Kar, Yuanzhen Li, Deqing Sun, Jonathan~T Barron, Hendrik Lensch, and Varun Jampani.
\newblock Samurai: Shape and material from unconstrained real-world arbitrary image collections.
\newblock {\em arXiv preprint arXiv:2205.15768}, 2022.

\bibitem{boss2021neural}
Mark Boss, Varun Jampani, Raphael Braun, Ce Liu, Jonathan Barron, and Hendrik Lensch.
\newblock Neural-pil: Neural pre-integrated lighting for reflectance decomposition.
\newblock {\em Advances in Neural Information Processing Systems}, 34:10691--10704, 2021.

\bibitem{cai2022neural}
Hongrui Cai, Wanquan Feng, Xuetao Feng, Yan Wang, and Juyong Zhang.
\newblock Neural surface reconstruction of dynamic scenes with monocular rgb-d camera.
\newblock {\em arXiv preprint arXiv:2206.15258}, 2022.

\bibitem{carranza2003free}
Joel Carranza, Christian Theobalt, Marcus~A Magnor, and Hans-Peter Seidel.
\newblock Free-viewpoint video of human actors.
\newblock {\em ACM transactions on graphics (TOG)}, 22(3):569--577, 2003.

\bibitem{chen2021snarf}
Xu Chen, Yufeng Zheng, Michael~J Black, Otmar Hilliges, and Andreas Geiger.
\newblock Snarf: Differentiable forward skinning for animating non-rigid neural implicit shapes.
\newblock In {\em Proceedings of the IEEE/CVF International Conference on Computer Vision}, pages 11594--11604, 2021.

\bibitem{chen2022tracing}
Ziyu Chen, Chenjing Ding, Jianfei Guo, Dongliang Wang, Yikang Li, Xuan Xiao, Wei Wu, and Li Song.
\newblock L-tracing: Fast light visibility estimation on neural surfaces by sphere tracing.
\newblock In {\em European Conference on Computer Vision}, pages 217--233. Springer, 2022.

\bibitem{chen2022relighting4d}
Zhaoxi Chen and Ziwei Liu.
\newblock Relighting4d: Neural relightable human from videos.
\newblock In {\em European Conference on Computer Vision}, pages 606--623. Springer, 2022.

\bibitem{collet2015high}
Alvaro Collet, Ming Chuang, Pat Sweeney, Don Gillett, Dennis Evseev, David Calabrese, Hugues Hoppe, Adam Kirk, and Steve Sullivan.
\newblock High-quality streamable free-viewpoint video.
\newblock {\em ACM Transactions on Graphics (ToG)}, 34(4):1--13, 2015.

\bibitem{blender2018}
Blender~Online Community.
\newblock {\em Blender - a 3D modelling and rendering package}.
\newblock Blender Foundation, Stichting Blender Foundation, Amsterdam, 2018.

\bibitem{debevec2008rendering}
Paul Debevec.
\newblock Rendering synthetic objects into real scenes: Bridging traditional and image-based graphics with global illumination and high dynamic range photography.
\newblock In {\em Acm siggraph 2008 classes}, pages 1--10. 2008.

\bibitem{debevec2012light}
Paul Debevec.
\newblock The light stages and their applications to photoreal digital actors.
\newblock Technical report, UNIVERSITY OF SOUTHERN CALIFORNIA LOS ANGELES, 2012.

\bibitem{debevec2000acquiring}
Paul Debevec, Tim Hawkins, Chris Tchou, Haarm-Pieter Duiker, Westley Sarokin, and Mark Sagar.
\newblock Acquiring the reflectance field of a human face.
\newblock In {\em Proceedings of the 27th annual conference on Computer graphics and interactive techniques}, pages 145--156, 2000.

\bibitem{fang2021mirrored}
Qi Fang, Qing Shuai, Junting Dong, Hujun Bao, and Xiaowei Zhou.
\newblock Reconstructing 3d human pose by watching humans in the mirror.
\newblock In {\em CVPR}, 2021.

\bibitem{grau2003studio}
Oliver Grau.
\newblock Studio production system for dynamic 3d content.
\newblock In {\em Visual Communications and Image Processing 2003}, volume 5150, pages 80--89. SPIE, 2003.

\bibitem{gropp2020implicit}
Amos Gropp, Lior Yariv, Niv Haim, Matan Atzmon, and Yaron Lipman.
\newblock Implicit geometric regularization for learning shapes.
\newblock {\em arXiv preprint arXiv:2002.10099}, 2020.

\bibitem{guo2019relightables}
Kaiwen Guo, Peter Lincoln, Philip Davidson, Jay Busch, Xueming Yu, Matt Whalen, Geoff Harvey, Sergio Orts-Escolano, Rohit Pandey, Jason Dourgarian, et~al.
\newblock The relightables: Volumetric performance capture of humans with realistic relighting.
\newblock {\em ACM Transactions on Graphics (ToG)}, 38(6):1--19, 2019.

\bibitem{habermann2019livecap}
Marc Habermann, Weipeng Xu, Michael Zollhoefer, Gerard Pons-Moll, and Christian Theobalt.
\newblock Livecap: Real-time human performance capture from monocular video.
\newblock {\em ACM Transactions On Graphics (TOG)}, 38(2):1--17, 2019.

\bibitem{habermann2020deepcap}
Marc Habermann, Weipeng Xu, Michael Zollhofer, Gerard Pons-Moll, and Christian Theobalt.
\newblock Deepcap: Monocular human performance capture using weak supervision.
\newblock In {\em Proceedings of the IEEE/CVF Conference on Computer Vision and Pattern Recognition}, pages 5052--5063, 2020.

\bibitem{hart1996sphere}
John~C Hart.
\newblock Sphere tracing: A geometric method for the antialiased ray tracing of implicit surfaces.
\newblock {\em The Visual Computer}, 12(10):527--545, 1996.

\bibitem{hart1993sphere}
John~C Hart et~al.
\newblock Sphere tracing: Simple robust antialiased rendering of distance-based implicit surfaces.
\newblock In {\em Siggraph}, volume~93, pages 1--11, 1993.

\bibitem{huang2020arch}
Zeng Huang, Yuanlu Xu, Christoph Lassner, Hao Li, and Tony Tung.
\newblock Arch: Animatable reconstruction of clothed humans.
\newblock In {\em Proceedings of the IEEE/CVF Conference on Computer Vision and Pattern Recognition}, pages 3093--3102, 2020.

\bibitem{iqbal2022rana}
Umar Iqbal, Akin Caliskan, Koki Nagano, Sameh Khamis, Pavlo Molchanov, and Jan Kautz.
\newblock Rana: Relightable articulated neural avatars.
\newblock {\em arXiv preprint arXiv:2212.03237}, 2022.

\bibitem{iwase2023relightablehands}
Shun Iwase, Shunsuke Saito, Tomas Simon, Stephen Lombardi, Timur Bagautdinov, Rohan Joshi, Fabian Prada, Takaaki Shiratori, Yaser Sheikh, and Jason Saragih.
\newblock Relightablehands: Efficient neural relighting of articulated hand models.
\newblock In {\em Proceedings of the IEEE/CVF Conference on Computer Vision and Pattern Recognition}, pages 16663--16673, 2023.

\bibitem{ji2022geometry}
Chaonan Ji, Tao Yu, Kaiwen Guo, Jingxin Liu, and Yebin Liu.
\newblock Geometry-aware single-image full-body human relighting.
\newblock In {\em European Conference on Computer Vision}, pages 388--405. Springer, 2022.

\bibitem{jiang2022neuman}
Wei Jiang, Kwang~Moo Yi, Golnoosh Samei, Oncel Tuzel, and Anurag Ranjan.
\newblock Neuman: Neural human radiance field from a single video.
\newblock In {\em European Conference on Computer Vision}, pages 402--418. Springer, 2022.

\bibitem{kajiya1986rendering}
James~T Kajiya.
\newblock The rendering equation.
\newblock In {\em Proceedings of the 13th annual conference on Computer graphics and interactive techniques}, pages 143--150, 1986.

\bibitem{kanade1997virtualized}
Takeo Kanade, Peter Rander, and PJ Narayanan.
\newblock Virtualized reality: Constructing virtual worlds from real scenes.
\newblock {\em IEEE multimedia}, 4(1):34--47, 1997.

\bibitem{kanamori2019relighting}
Yoshihiro Kanamori and Yuki Endo.
\newblock Relighting humans: occlusion-aware inverse rendering for full-body human images.
\newblock {\em arXiv preprint arXiv:1908.02714}, 2019.

\bibitem{kingma2014adam}
Diederik~P Kingma and Jimmy Ba.
\newblock Adam: A method for stochastic optimization.
\newblock {\em arXiv preprint arXiv:1412.6980}, 2014.

\bibitem{kuang2022neroic}
Zhengfei Kuang, Kyle Olszewski, Menglei Chai, Zeng Huang, Panos Achlioptas, and Sergey Tulyakov.
\newblock Neroic: Neural rendering of objects from online image collections.
\newblock {\em arXiv preprint arXiv:2201.02533}, 2022.

\bibitem{lewis2000pose}
John~P Lewis, Matt Cordner, and Nickson Fong.
\newblock Pose space deformation: a unified approach to shape interpolation and skeleton-driven deformation.
\newblock In {\em Proceedings of the 27th annual conference on Computer graphics and interactive techniques}, pages 165--172, 2000.

\bibitem{li2023megane}
Junxuan Li, Shunsuke Saito, Tomas Simon, Stephen Lombardi, Hongdong Li, and Jason Saragih.
\newblock Megane: Morphable eyeglass and avatar network.
\newblock In {\em Proceedings of the IEEE/CVF Conference on Computer Vision and Pattern Recognition}, pages 12769--12779, 2023.

\bibitem{li2021ai}
Ruilong Li, Shan Yang, David~A Ross, and Angjoo Kanazawa.
\newblock Ai choreographer: Music conditioned 3d dance generation with aist++.
\newblock In {\em Proceedings of the IEEE/CVF International Conference on Computer Vision}, pages 13401--13412, 2021.

\bibitem{lin2022robust}
Shanchuan Lin, Linjie Yang, Imran Saleemi, and Soumyadip Sengupta.
\newblock Robust high-resolution video matting with temporal guidance.
\newblock In {\em Proceedings of the IEEE/CVF Winter Conference on Applications of Computer Vision}, pages 238--247, 2022.

\bibitem{liu2021neural}
Lingjie Liu, Marc Habermann, Viktor Rudnev, Kripasindhu Sarkar, Jiatao Gu, and Christian Theobalt.
\newblock Neural actor: Neural free-view synthesis of human actors with pose control.
\newblock {\em ACM Transactions on Graphics (TOG)}, 40(6):1--16, 2021.

\bibitem{loper2015smpl}
Matthew Loper, Naureen Mahmood, Javier Romero, Gerard Pons-Moll, and Michael~J Black.
\newblock Smpl: A skinned multi-person linear model.
\newblock {\em ACM transactions on graphics (TOG)}, 34(6):1--16, 2015.

\bibitem{meka2020deep}
Abhimitra Meka, Rohit Pandey, Christian Haene, Sergio Orts-Escolano, Peter Barnum, Philip David-Son, Daniel Erickson, Yinda Zhang, Jonathan Taylor, Sofien Bouaziz, et~al.
\newblock Deep relightable textures: volumetric performance capture with neural rendering.
\newblock {\em ACM Transactions on Graphics (TOG)}, 39(6):1--21, 2020.

\bibitem{mildenhall2021nerf}
Ben Mildenhall, Pratul~P Srinivasan, Matthew Tancik, Jonathan~T Barron, Ravi Ramamoorthi, and Ren Ng.
\newblock Nerf: Representing scenes as neural radiance fields for view synthesis.
\newblock {\em Communications of the ACM}, 65(1):99--106, 2021.

\bibitem{pandey2021total}
Rohit Pandey, Sergio~Orts Escolano, Chloe Legendre, Christian Haene, Sofien Bouaziz, Christoph Rhemann, Paul Debevec, and Sean Fanello.
\newblock Total relighting: learning to relight portraits for background replacement.
\newblock {\em ACM Transactions on Graphics (TOG)}, 40(4):1--21, 2021.

\bibitem{parker1998single}
Steven Parker, Peter Shirley, and Brian Smits.
\newblock Single sample soft shadows.
\newblock Technical report, Technical Report UUCS-98-019, Computer Science Department, University of Utah, 1998.

\bibitem{peng2021animatable}
Sida Peng, Junting Dong, Qianqian Wang, Shangzhan Zhang, Qing Shuai, Xiaowei Zhou, and Hujun Bao.
\newblock Animatable neural radiance fields for modeling dynamic human bodies.
\newblock In {\em Proceedings of the IEEE/CVF International Conference on Computer Vision}, pages 14314--14323, 2021.

\bibitem{peng2022animatable}
Sida Peng, Zhen Xu, Junting Dong, Qianqian Wang, Shangzhan Zhang, Qing Shuai, Hujun Bao, and Xiaowei Zhou.
\newblock Animatable implicit neural representations for creating realistic avatars from videos.
\newblock {\em arXiv preprint arXiv:2203.08133}, 2022.

\bibitem{peng2021neural}
Sida Peng, Yuanqing Zhang, Yinghao Xu, Qianqian Wang, Qing Shuai, Hujun Bao, and Xiaowei Zhou.
\newblock Neural body: Implicit neural representations with structured latent codes for novel view synthesis of dynamic humans.
\newblock In {\em Proceedings of the IEEE/CVF Conference on Computer Vision and Pattern Recognition}, pages 9054--9063, 2021.

\bibitem{romero2017embodied}
Javier Romero, Dimitrios Tzionas, and Michael~J Black.
\newblock Embodied hands: Modeling and capturing hands and bodies together.
\newblock {\em ACM ToG}, 2017.

\bibitem{roussopoulos1995nearest}
Nick Roussopoulos, Stephen Kelley, and Frederic Vincent.
\newblock Nearest neighbor queries.
\newblock In {\em Proceedings of the 1995 ACM SIGMOD international conference on Management of data}, pages 71--79, 1995.

\bibitem{saito2021scanimate}
Shunsuke Saito, Jinlong Yang, Qianli Ma, and Michael~J Black.
\newblock Scanimate: Weakly supervised learning of skinned clothed avatar networks.
\newblock In {\em Proceedings of the IEEE/CVF Conference on Computer Vision and Pattern Recognition}, pages 2886--2897, 2021.

\bibitem{schmitt2020joint}
Carolin Schmitt, Simon Donne, Gernot Riegler, Vladlen Koltun, and Andreas Geiger.
\newblock On joint estimation of pose, geometry and svbrdf from a handheld scanner.
\newblock In {\em Proceedings of the IEEE/CVF Conference on Computer Vision and Pattern Recognition}, pages 3493--3503, 2020.

\bibitem{schonberger2016structure}
Johannes~L Schonberger and Jan-Michael Frahm.
\newblock Structure-from-motion revisited.
\newblock In {\em Proceedings of the IEEE conference on computer vision and pattern recognition}, pages 4104--4113, 2016.

\bibitem{seyb2019non}
Dario Seyb, Alec Jacobson, Derek Nowrouzezahrai, and Wojciech Jarosz.
\newblock Non-linear sphere tracing for rendering deformed signed distance fields.
\newblock {\em ACM Transactions on Graphics}, 38(6), 2019.

\bibitem{shapiro2014rapid}
Ari Shapiro, Andrew Feng, Ruizhe Wang, Hao Li, Mark Bolas, Gerard Medioni, and Evan Suma.
\newblock Rapid avatar capture and simulation using commodity depth sensors.
\newblock {\em Computer Animation and Virtual Worlds}, 25(3-4):201--211, 2014.

\bibitem{sharp2022spelunking}
Nicholas Sharp and Alec Jacobson.
\newblock Spelunking the deep: guaranteed queries on general neural implicit surfaces via range analysis.
\newblock {\em ACM Transactions on Graphics (TOG)}, 41(4):1--16, 2022.

\bibitem{srinivasan2021nerv}
Pratul~P Srinivasan, Boyang Deng, Xiuming Zhang, Matthew Tancik, Ben Mildenhall, and Jonathan~T Barron.
\newblock Nerv: Neural reflectance and visibility fields for relighting and view synthesis.
\newblock In {\em Proceedings of the IEEE/CVF Conference on Computer Vision and Pattern Recognition}, pages 7495--7504, 2021.

\bibitem{starck2005virtual}
Jonathan Starck and Adrian Hilton.
\newblock Virtual view synthesis of people from multiple view video sequences.
\newblock {\em Graphical Models}, 67(6):600--620, 2005.

\bibitem{starck2007surface}
Jonathan Starck and Adrian Hilton.
\newblock Surface capture for performance-based animation.
\newblock {\em IEEE computer graphics and applications}, 27(3):21--31, 2007.

\bibitem{tong2012scanning}
Jing Tong, Jin Zhou, Ligang Liu, Zhigeng Pan, and Hao Yan.
\newblock Scanning 3d full human bodies using kinects.
\newblock {\em IEEE transactions on visualization and computer graphics}, 18(4):643--650, 2012.

\bibitem{vlasic2008articulated}
Daniel Vlasic, Ilya Baran, Wojciech Matusik, and Jovan Popovi{\'c}.
\newblock Articulated mesh animation from multi-view silhouettes.
\newblock In {\em Acm Siggraph 2008 papers}, pages 1--9. 2008.

\bibitem{walter2007microfacet}
Bruce Walter, Stephen~R Marschner, Hongsong Li, and Kenneth~E Torrance.
\newblock Microfacet models for refraction through rough surfaces.
\newblock In {\em Proceedings of the 18th Eurographics conference on Rendering Techniques}, pages 195--206, 2007.

\bibitem{wang2022arah}
Shaofei Wang, Katja Schwarz, Andreas Geiger, and Siyu Tang.
\newblock Arah: Animatable volume rendering of articulated human sdfs.
\newblock In {\em European conference on computer vision}, pages 1--19. Springer, 2022.

\bibitem{weng2022humannerf}
Chung-Yi Weng, Brian Curless, Pratul~P Srinivasan, Jonathan~T Barron, and Ira Kemelmacher-Shlizerman.
\newblock Humannerf: Free-viewpoint rendering of moving people from monocular video.
\newblock {\em arXiv preprint arXiv:2201.04127}, 2022.

\bibitem{wenger2005performance}
Andreas Wenger, Andrew Gardner, Chris Tchou, Jonas Unger, Tim Hawkins, and Paul Debevec.
\newblock Performance relighting and reflectance transformation with time-multiplexed illumination.
\newblock {\em ACM Transactions on Graphics (TOG)}, 24(3):756--764, 2005.

\bibitem{xu2021h}
Hongyi Xu, Thiemo Alldieck, and Cristian Sminchisescu.
\newblock H-nerf: Neural radiance fields for rendering and temporal reconstruction of humans in motion.
\newblock {\em Advances in Neural Information Processing Systems}, 34:14955--14966, 2021.

\bibitem{xu2018monoperfcap}
Weipeng Xu, Avishek Chatterjee, Michael Zollh{\"o}fer, Helge Rhodin, Dushyant Mehta, Hans-Peter Seidel, and Christian Theobalt.
\newblock Monoperfcap: Human performance capture from monocular video.
\newblock {\em ACM Transactions on Graphics (ToG)}, 37(2):1--15, 2018.

\bibitem{yariv2021volume}
Lior Yariv, Jiatao Gu, Yoni Kasten, and Yaron Lipman.
\newblock Volume rendering of neural implicit surfaces.
\newblock {\em Advances in Neural Information Processing Systems}, 34:4805--4815, 2021.

\bibitem{yeh2022learning}
Yu-Ying Yeh, Koki Nagano, Sameh Khamis, Jan Kautz, Ming-Yu Liu, and Ting-Chun Wang.
\newblock Learning to relight portrait images via a virtual light stage and synthetic-to-real adaptation.
\newblock {\em ACM Transactions on Graphics (TOG)}, 41(6):1--21, 2022.

\bibitem{zhang2021physg}
Kai Zhang, Fujun Luan, Qianqian Wang, Kavita Bala, and Noah Snavely.
\newblock Physg: Inverse rendering with spherical gaussians for physics-based material editing and relighting.
\newblock In {\em Proceedings of the IEEE/CVF Conference on Computer Vision and Pattern Recognition}, pages 5453--5462, 2021.

\bibitem{zhang2018unreasonable}
Richard Zhang, Phillip Isola, Alexei~A Efros, Eli Shechtman, and Oliver Wang.
\newblock The unreasonable effectiveness of deep features as a perceptual metric.
\newblock In {\em Proceedings of the IEEE conference on computer vision and pattern recognition}, pages 586--595, 2018.

\bibitem{zhang2021neural}
Xiuming Zhang, Sean Fanello, Yun-Ta Tsai, Tiancheng Sun, Tianfan Xue, Rohit Pandey, Sergio Orts-Escolano, Philip Davidson, Christoph Rhemann, Paul Debevec, et~al.
\newblock Neural light transport for relighting and view synthesis.
\newblock {\em ACM Transactions on Graphics (TOG)}, 40(1):1--17, 2021.

\bibitem{zhang2021nerfactor}
Xiuming Zhang, Pratul~P Srinivasan, Boyang Deng, Paul Debevec, William~T Freeman, and Jonathan~T Barron.
\newblock Nerfactor: Neural factorization of shape and reflectance under an unknown illumination.
\newblock {\em ACM Transactions on Graphics (TOG)}, 40(6):1--18, 2021.

\bibitem{zhang2022modeling}
Yuanqing Zhang, Jiaming Sun, Xingyi He, Huan Fu, Rongfei Jia, and Xiaowei Zhou.
\newblock Modeling indirect illumination for inverse rendering.
\newblock In {\em Proceedings of the IEEE/CVF Conference on Computer Vision and Pattern Recognition}, pages 18643--18652, 2022.

\bibitem{zheng2022structured}
Zerong Zheng, Han Huang, Tao Yu, Hongwen Zhang, Yandong Guo, and Yebin Liu.
\newblock Structured local radiance fields for human avatar modeling.
\newblock In {\em Proceedings of the IEEE/CVF Conference on Computer Vision and Pattern Recognition}, pages 15893--15903, 2022.

\end{thebibliography}
